\documentclass{llncs}
\usepackage[T1]{fontenc}
\usepackage{graphicx}
\usepackage{subfig}
\usepackage{bbm}
\usepackage{color}
\usepackage{amsbsy}
\usepackage{amsmath}
\usepackage{amsfonts}
\usepackage{hyperref}
\usepackage{amssymb}
\usepackage{algorithmic}
\usepackage[misc]{ifsym}
\usepackage{algorithm}
\usepackage{float}
\usepackage[flushleft]{threeparttable}
\usepackage{booktabs}
\usepackage{xcolor}
\usepackage[absolute,overlay]{textpos}
\newcommand\mycommfont[1]{\small\ttfamily\textcolor{blue}{#1}}

\begin{document}
\begin{figure}[t]
\hspace{-3.6cm}
  \includegraphics[width=1.6\textwidth]{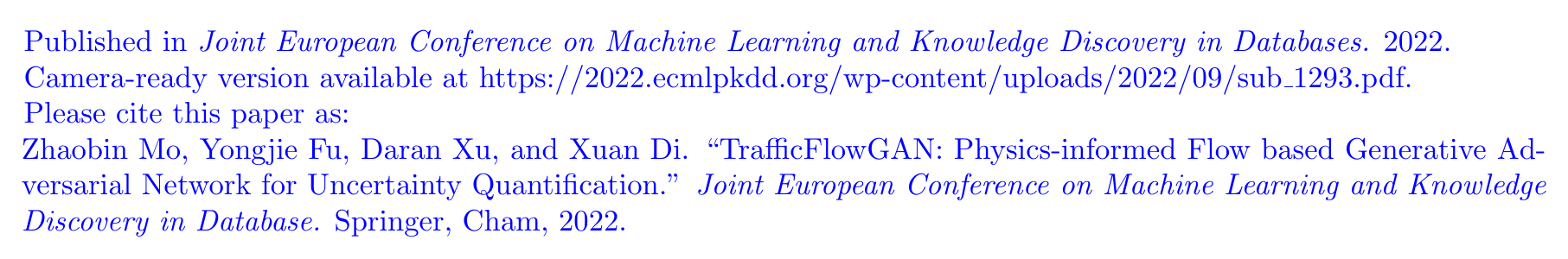}
\end{figure}

\title{
TrafficFlowGAN: Physics-informed Flow based Generative Adversarial Network
for Uncertainty Quantification}
\titlerunning{TrafficFlowGAN for Uncertainty Quantification}
%
\author{Zhaobin Mo\inst{1}, Yongjie Fu\inst{1}, Daran Xu\inst{2}, and Xuan Di\inst{1,3}}
\institute{Department of Civil Engineering and Engineering Mechanics, Columbia University, New York, USA 
\\
\email{\{zm2302,yf2578,xd2187\}@columbia.edu}
\and
Department of Statistics, Columbia University, New York, USA\\
\email{dx2207@columbia.edu}
\and
Data Science Institute, Columbia University, New York, USA}
\authorrunning{Z. Mo et al.}
%
%
\maketitle              
\begin{abstract}
This paper proposes the TrafficFlowGAN, a physics-informed flow based generative adversarial network (GAN), for uncertainty quantification (UQ) of dynamical systems. TrafficFlowGAN adopts a normalizing flow model as the generator to explicitly estimate the data likelihood. This flow model is trained to maximize the data likelihood and to generate synthetic data that can fool a convolutional discriminator. We further regularize this training process using prior physics information, so-called physics-informed deep learning (PIDL). To the best of our knowledge, we are the first to propose an integration of normalizing flow, GAN and PIDL for the UQ problems. We take the traffic state estimation (TSE), which aims to estimate the traffic variables (e.g. traffic density and velocity) using partially observed data, as an example to demonstrate the performance of our proposed model. We conduct numerical experiments where the proposed model is applied to learn the solutions of stochastic differential equations. The results demonstrate the robustness and accuracy of the proposed model, together with the ability to learn a machine learning surrogate model. We also test it on a real-world dataset, the Next Generation SIMulation (NGSIM), to show that the proposed TrafficFlowGAN can outperform the baselines, including the pure flow model, the physics-informed flow model, and the flow based GAN model. Source code and data are available at \href{https://github.com/ZhaobinMo/TrafficFlowGAN}{https://github.com/ZhaobinMo/TrafficFlowGAN}.

\keywords{Uncertainty Quantification (UQ)  \and Normalizing Flow \and Generative Adversarial Networks (GAN) \and Physics-informed Deep Learning (PIDL).}
\end{abstract}
\section{Introduction}
Uncertainty quantification (UQ) is the process of characterizing the uncertainty of system dynamics, accounting for two main sources of uncertainty \cite{national2012assessing}. 
The \emph{aleatoric uncertainty} (or data uncertainty) refers to uncertainty arising from external factors, such as measurement noise, and random initial or boundary conditions. The \emph{epistemic uncertainty} arises from the inadequate knowledge of the underlying model, such as inherent stochasticity in human behavior. With these random factors, UQ of dynamical systems is crucial to avoid potential system oscillation or cascading errors. 

Two classes of methods are developed to characterize the aforementioned sources of uncertainty. 
The \textbf{physics-based} method assumes that the observations are generated from underlying physics imposed by Gaussian noise; thus filtering methods or Bayesian inference can be applied to propagate uncertainty. However, the physics-based method suffers from limitations such as non-Gaussian likelihoods and high-dimensional posterior distributions \cite{yang2019adversarial}. In contrast, the \textbf{data-driven} method, such as generative adversarial networks (GAN) \cite{goodfellow2014generative}, tries to characterize any distribution of data directly without making any assumption of noise. 
Recently, there is a growing trend in integrating physics-based models into the data-driven framework, namely, \textbf{physics-informed deep learning} (PIDL) \cite{Raissi-2018a}. PIDL-based UQ methods can characterize generic data distribution while ensuring physics consistency.

Among all PIDL models for UQ, the physics-informed GAN is the most widely used, which has been applied to solve stochastic differential equations \cite{yang2020physics,yang2019adversarial,daw2021pid} and quantify uncertainty in various domains \cite{siddani2021machine,shi2021physics}. Although GAN generates high-quality samples \cite{grover2018flow} through adversarial training, it has stability and convergence issues. Moreover, as GAN cannot calculate the model likelihood, it may miss important modes of the data distribution, namely, \emph{mode collapse} \cite{theis2015note}. In contrast, normalizing flow \cite{dinh2016density} calculates the exact data likelihood and is trained using maximum likelihood estimation (MLE), which is an effective way to avoid mode collapse. However, applying PIDL to the normalizing flow is still at its nascent stage and we only find one relevant work \cite{guo2021normalizing}.  

Leveraging the pros of both the MLE and adversarial training, Flow-GAN, a combination of normalizing flow and GAN, is first introduced in \cite{grover2018flow}, which can achieve both high data likelihood and good sample qualities. Flow-GAN has been applied to 
manifold learning \cite{brehmer2020flows} and image-to-image translation \cite{grover2020alignflow}. Little research has been documented that applies Flow-GAN to UQ problems.

In this paper, we propose TrafficFlowGAN that leverages likelihood training, adversarial training, and PIDL for the UQ problems. To the best of our knowledge, we are the first to integrate these three methods for the UQ problems. Main contributions of this paper include:
\begin{itemize}
    \item We propose a hybrid generative model, TrafficFlowGAN, combining normalizing flow and GAN to achieve both high likelihoods and good sample qualities and to avoid mode collapse. 
    \item We incorporate physics information into the TrafficFlowGAN model for estimation accuracy and data efficiency, and use neural network surrogate models to learn the inter-relations between the physics variables at the same time. 
    \item We apply the TrafficFlowGAN model to learn solutions of second-order stochastic partial differential equations  (PDEs), and demonstrate the performance of TrafficFlowGAN by applying it to a traffic state estimation (TSE) problem with real-world data. 
\end{itemize}
The rest of this paper is organized as follows: Section \ref{sec:background} introduces the background and related work. Section \ref{sec:method} introduces the structure of TrafficFlowGAN for the UQ problems. Section \ref{sec:experiment} demonstrates how TrafficFlowGAN learns solutions of a PDE and the relations between physics variables using a neural network surrogate model. Section \ref{sec:case_study} demonstrates how TrafficFlowGAN characterizes uncertainty from the real-world data in the TSE problem, where two traffic models, i.e. the Aw–Rascle–Zhang (ARZ) \cite{Aw-Rascle-2000} and the Lighthill-Whitham-Richards (LWR) \cite{Lighthill-1955} models, are used as the physics components. Section \ref{sec:conclusion} concludes our work and projects future directions in this promising arena.

\section{Background and Related Work}\label{sec:background}

\subsection{Normalizing Flow}\label{sec:flow_framework}

The flow model aims to learn an invertible function $\pmb{z}=f_{\theta}(\pmb{u}): \mathbb{R}^{D} \mapsto \mathbb{R}^{D}$, where data $\pmb{u}$ is sampled from a distribution $ p_{\text{data}}(\pmb{u})$ and $\pmb{z} \sim p_{\pmb{z}}(\pmb{z})$ is a random noise of the same dimension as the data. The data likelihood $p_{\theta}(\pmb{u})$ can be explicitly expressed by using the \emph{change of variable formula}:

\begin{equation}\label{eqn:change_variable}
     p_{\theta}(\pmb{u})=p_{\pmb{z}}(\pmb{z})\left|\operatorname{det}\left(\frac{\partial f_{\theta}^{-1}(\pmb{z})}{\partial \pmb{z}}\right)\right|^{-1}.
\end{equation}
To compute $p_{\theta}(\pmb{u})$, it is nontrivial to choose a latent variable $\pmb{z}$ that has an easy form and to design the invertible function $f_{\theta}$ so that the Jacobian determinant can be easily computed. A common selection of the latent variable $\pmb{z}$ is the standard Gaussian, i.e. $p_{\mathcal{\pmb{z}}}(\pmb{z}) {\sim} \mathcal{N}(0,\pmb{I}_D)$. To compute the Jacobian determinant, RealNVP \cite{dinh2016density} designs the invertible function $f_{\theta}$ as an \emph{affine coupling transformation} following the equations below:

\begin{equation}\label{eqn:flow_forward}
\begin{aligned}
    f_{\theta} : 
       = 
        \begin{cases}
        \pmb{z}_{1:d} = \pmb{u}_{1:d} \\
    \pmb{z}_{d+1:D} = \pmb{u}_{d+1:D} \odot e^{k_{\theta}(\pmb{u}_{1:d})} + b_{\theta}(\pmb{u}_{1:d})
        \end{cases}
\end{aligned},
\end{equation}
where $\pmb{u}$ and $\pmb{z}$ are split into two partitions at the $d$th elements. The \emph{scale function} $k_{\theta}$ and the \emph{translation function} $b_{\theta}$ are neural networks to be learned, which constitute the affine transformation of the partition $\pmb{u}_{1:d}$. $\odot$ is the Hadamard product or element-wise product. By this design of invertible function, the Jacobian determinant in 
Eq.~\ref{eqn:change_variable} can be computed by

\begin{equation} \label{eqn:jacobian}
    \left|\operatorname{det}\left(\frac{\partial f_{\theta}^{-1}(\pmb{z})}{\partial \pmb{z}}\right)\right|^{-1} = e ^ { \sum_j \left[ k_{\theta}(\pmb{z}_{1:d})\right] _j },
\end{equation}
where $j$ is the index of the element of $k_{\theta}(\pmb{z}_{1:d})$. The inverse function $f^{-1}_{\theta}$ can also be obtained by
\begin{equation}\label{eqn:flow_inverse}
    f_{\theta}^{-1} : 
       = 
    \begin{cases}
    \pmb{u}_{1:d} = \pmb{z}_{1:d} \\
    \pmb{u}_{d+1:D}  = (\pmb{z}_{d+1:D} - b_{\theta}(\pmb{z}_{1:d})) / e^{k_{\theta}(\pmb{z}_{1:d})}
    \end{cases}.
\end{equation}

To better accommodate the complex data distribution, $f_{\theta}$ is further modeled as a sequence of affine coupling transformations: $f_{\theta} = f_L \circ ... \circ f_1$, where $L$ is the total number of transformations. Let $f_{l}$ be the $l$th invertible mapping and $\pmb{h}^{(l)}$ be the $l$th latent variable that satisfies $\pmb{h}^{(l)}=f_{l}(\pmb{h}^{(l-1)})$, where $\pmb{h}^{(0)}=\pmb{u}$ and $\pmb{h}^{(L)} = \pmb{z}$. Then the log-likelihood of $\pmb{u}$ can be computed by:
\begin{equation} \label{eqn:flow_coupling}
    \begin{aligned}
    \log p_{\theta}(\pmb{u}) 
    & = \log p_{\pmb{z}}(\pmb{z}) + \sum_{l=0}^{L-1} \log \left|\operatorname{det}\left(\frac{\partial f_{l}^{-1}(\pmb{h}^{(l)})}{\partial \pmb{h}^{(l)}}\right)\right|^{-1}.
    \end{aligned}
\end{equation}
The computation of the log-likelihood of $\pmb{z}$ is straightforward as $\pmb{z}$ is assumed to follow a standard Gaussian distribution, and each Jacobian determinant can be calculated following Eq.~\ref{eqn:jacobian}. Thus, the exact data likelihood is tractable and the flow model can be trained by the MLE.

\subsection{Generative Adversarial Network (GAN)}
GAN aims to train a generator $G_{\theta}$ to learn the mapping from a random noise $\pmb{z}$ to the corresponding state variables $\pmb{u}$, i.e. $G_{\theta}: \pmb{z} \rightarrow \pmb{u}$. The objective of the generator $G_{\theta}$ is to fool an adversarially trained discriminator $D_{\phi}$. Different GAN variants use different metrics to evaluate the divergence between the prediction distribution and the data distribution, such as the Kullback-Leibler (KL) divergence, the Jensen-Shannon divergence, and the Wasserstein distance \cite{arjovsky2017wasserstein}. Among these metrics, the Wasserstein distance has received growing popularity for its stability, which optimizes the following objective: 
\begin{equation} \label{eqn:wgan}
    \begin{aligned}
    \min _{\theta} \max_{\phi \in \mathcal{F}} \mathbb{E}_{ p_{\text{data}}(\pmb{u})} \left[  D_{\phi}(\pmb{u}))\right] 
    -\mathbb{E}_{p_{\pmb{z}}(\pmb{z})}\left[D_{\phi}(G_{\theta}(\pmb{z}))\right],
    \end{aligned}
\end{equation}
where $\theta$ and $\phi$ are the parameters of the generator and the discriminator, respectively. $\mathcal{F}$ is defined such that $D_{\phi}$ is 1-Lipschitz.


\section{Framework of TrafficFlowGAN} \label{sec:method}

\subsection{Problem Statement}
Define the spatial and temporal domains as $\mathcal{X}$ and $\mathcal{T}$, respectively. $(x,t) \in \mathcal{X} \times \mathcal{T}$ is the spatio-temporal coordinate (``coordinate'' for short). It is assumed that the state variable $\pmb{u}$ can only be observed by limited number of sensors placed at fixed locations and at a specific frequency. Thus, we further define the \emph{observed (labeled) region} $O \subseteq \mathcal{X} \times \mathcal{T}$ as the spatio-temporal region where the state variable $\pmb{u}$ is observed, and thereby the \emph{unobserved (unlabeled) region} $C=\mathcal{X} \times \mathcal{T} \setminus O$. We represent the continuous domain in a discrete manner using grid points. Thus, the observed region $O$ and the unobserved region $C$ can be represented as collections of discrete coordinates: $O = \{(x_o^{(i)}, t_o^{(i)}) \}_{i=1}^{N_o}$ and $C = \{(x_c^{(j)}, t_c^{(j)})  \}_{j=1}^{Nc}$, where $i$ and $j$ are the indices of observed and unobserved coordinates, respectively; $N_o, N_c$ are the numbers of observed and unobserved coordinates, respectively. 



The state variable $\pmb{u}$ is a random variable for each coordinate, i.e. $\pmb{u} \sim p_{data}(\pmb{u}|x,t)$. Our goal is to train a generator such that its prediction distribution distribution $p_{\theta}(\hat{\pmb{u}}|x,t)$ matches the data distribution $p_{data}(\pmb{u}|x,t)$. Below we will introduce how to achieve this goal with our proposed TrafficFlowGAN.

\subsection{Overview of TrafficFlowGAN Structure}

\begin{figure}[htbp]
  \centering
  \includegraphics[width=\linewidth]{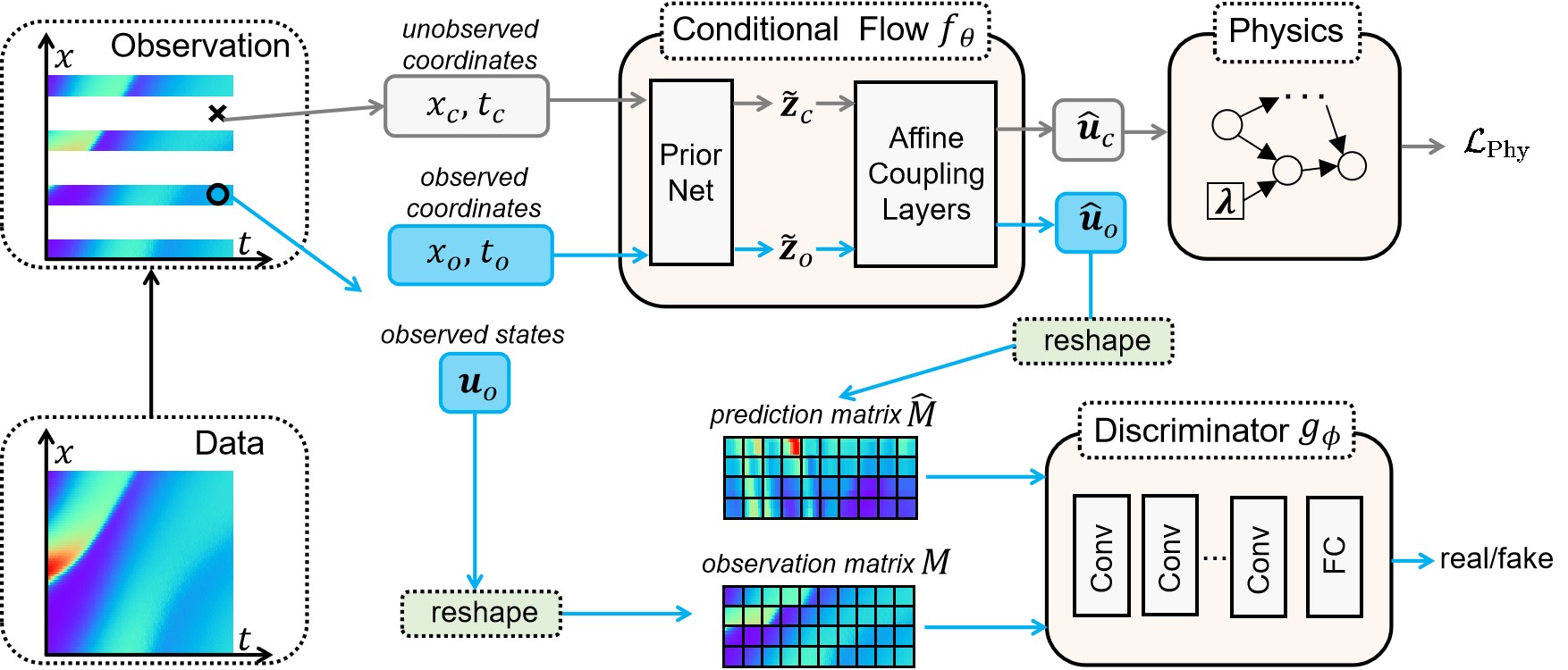}
  \caption{Structure of the TrafficFlowGAN.}
  \label{fig:physflowgan}
\end{figure}

 An overview of TrafficFlowGAN is illustrated in Fig.~\ref{fig:physflowgan}, which consists of three main components, namely, a conditional flow $f_{\theta}$, a physics-based computational graph, and a convolutional discriminator $D_{\phi}$. The data is illustrated as a heatmap in the spatio-temporal domain. We assume the data is measured by sensors at fixed locations. Due to limited range each sensor can cover, the observation region consists of separate horizontal ``strips.'' The observed and unobserved coordinates are fed into the conditional flow model to generate predictions $\hat{\pmb{u}}_o$ and $\hat{\pmb{u}}_c$, respectively. Those predictions bifurcate into two branches. In the upper branch, $\hat{\pmb{u}}_c$ are fed into a physics-based computational graph, which encodes physics laws, to calculate the physics loss function. This process of calculating physics loss from the unobserved coordinates is illustrated by grey arrows. In the lower branch, the prediction states $\hat{\pmb{u}}_o$ and the observed states $\pmb{u}_o$ are then reshaped to constitute the prediction matrix $\hat{M}$ and the observation matrix $M$, respectively. These two matrices are then fed into the convolutional discriminator. The process of calculating the adversarial loss from observed coordinates and states is illustrated by blue arrows. 

We will detail each component sequentially and explain how we integrate those components in the following subsections.  

\subsection{Convolutional Neural Network as the Discriminator} 
Existing physics-informed GAN models construct the discriminator as a \emph{fully connected network}. By explicitly adding the spatio-temporal coordinates into the input space, the discriminator is expected to make decisions based on the \emph{spatio-temporal pattern}, which will be represented better by the convolutional neural network (CNN). In this work, we propose to use CNN as the discriminator.

The discriminator $D_{\phi}$ consists of a sequence of convolutional layers (Conv) followed by a fully connected layer (FC). This FC layer outputs a $1 \times 1$ scalar indicating if the input matrix is from observations or predictions. The pooling layer is not used in this structure, as there is no requirement for compression in our task.

The reshape of the observation $\pmb{u}$ is straightforward. As we represent the spatio-temporal domain in a discrete manner, $\pmb{u}(x,t)$ for each coordinate can be viewed as a ``pixel'', and the dimension of $\pmb{u}$ is its number of channels. This is the same for reshaping the prediction $\hat{\pmb{u}}$ to get the prediction matrix $\hat{M}$. Note that due to randomness in data and predictions, we can sample multiple observation matrices $\{M^{(i)}\}_{i=1}^{N_{\omega}}$ and prediction matrices $\{\hat{M}^{(i)}\}_{i=1}^{N_{\omega}}$, where $N_{\omega}$ is the total number of sampling.

The discriminator $D_{\phi}$ can be updated by minimizing the Wasserstein loss:
\begin{equation} \label{eqn:loss_d}
    \mathcal{L}_{D}(\phi) = -\frac{1}{N_{\omega}} \sum_{i=1}^{N_{\omega}}  D_{\phi}(M^{(i)}) - D_{\phi}(\hat{M}^{(i)}) .
\end{equation}

\vspace{-5mm}
\subsection{Conditional Flow as the Generator}

\begin{figure}[htbp] 
  \centering
  \includegraphics[width=\linewidth]{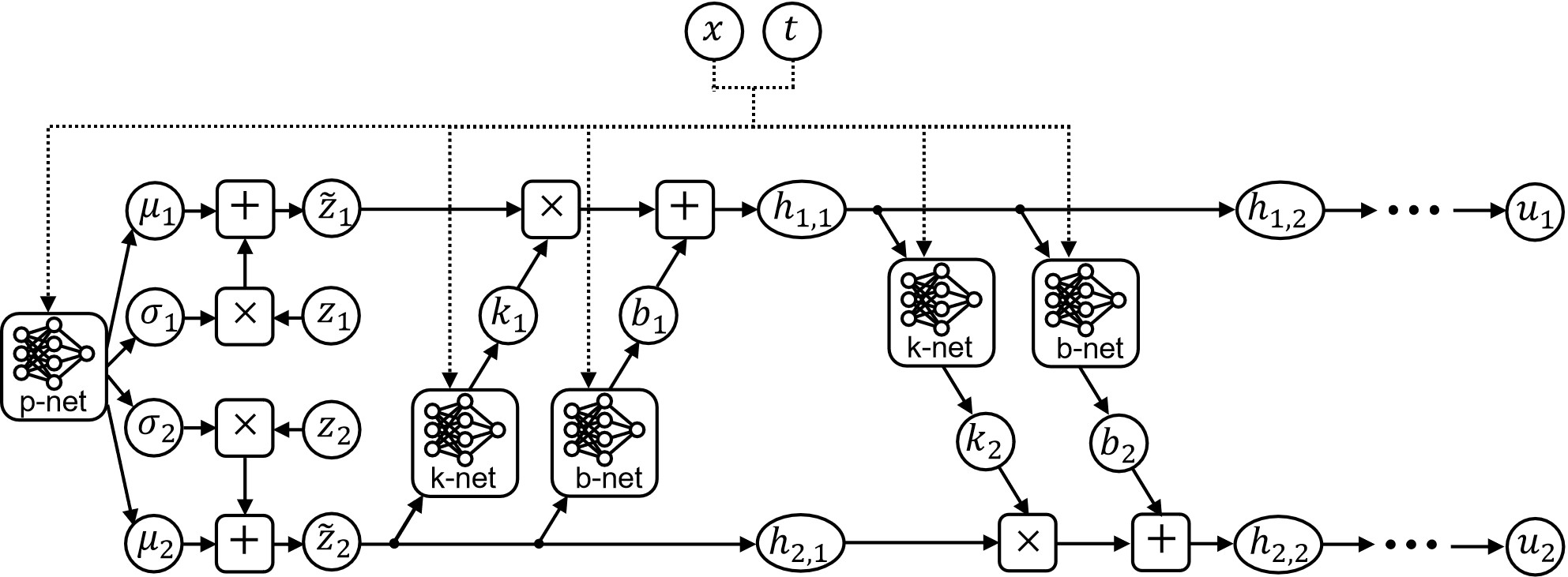}
  \caption{Structure of the conditional flow with a prior network.}
  \label{fig:cflow_pnet}
\end{figure}

We construct a conditional flow as our generator, as illustrated in Fig.~\ref{fig:cflow_pnet}. Assume $\pmb{u}$ has two elements, i.e. $u_1$ and $u_2$. Different from the tradition normalizing flow, we add a \emph{prior network} (p-net) to transform the standard Gaussian prior $\pmb{z}=(z_1, z_2)$ to $\tilde{\pmb{z}}=(\tilde{z}_1, \tilde{z}_2)$ with shifting and scaling, considering that the magnitude of uncertainty at different $(x,t)$ coordinates can be different. The prior network takes as input the coordinate $(x,t)$ and outputs the prior mean $\pmb{\mu} = (\mu_1, \mu_2)$ and prior standard deviation $\pmb{\sigma} = (\sigma_1, \sigma_2)$; thus $\tilde{\pmb{z}} \sim \mathcal{N}(\pmb{\mu}, \pmb{\sigma})$. The prior network is followed by affine coupling layers. Each affine coupling layer consists of a scale function (k-net) and a translation function (b-net), as introduced in Section~\ref{sec:flow_framework}.

Based on \cite{zang2020moflow} and our experiment, the exponential operation in Eq.~\ref{eqn:flow_forward} is numerically unstable, which may result in gradient explosion. Instead of using the RealNVP, we replace the exponential operation in Eq.~\ref{eqn:flow_forward} with a Sigmoid operation:
\begin{equation}\label{eqn:flow_forward_sigmoid}
\begin{aligned}
    f_{\theta} : 
       = 
        \begin{cases}
        \tilde{\pmb{z}}_{1:d} = \pmb{u}_{1:d} \\
    \tilde{\pmb{z}}_{d+1:D} = \pmb{u}_{d+1:D} \odot \text{Sigmoid}({k_{\theta}(\pmb{u}_{1:d};x,t)}) + b_{\theta}(\pmb{u}_{1:d};x,t)
        \end{cases}
\end{aligned},
\end{equation}
and the calculation of Jacobian determinant in Eq.~\ref{eqn:flow_coupling}  is thus changed to
\begin{equation} \label{eqn:jacobian_sigmoid}
    \left|\operatorname{det}\left(\frac{\partial f_{\theta}^{-1}(\tilde{\pmb{z}})}{\partial \tilde{\pmb{z}}}\right)\right|^{-1} =  { \sum_j \left[ k_{\theta}(\tilde{\pmb{z}}_{1:d};x,t)\right]_j },
\end{equation}
where $j$ is the element index of $k_{\theta}(\tilde{\pmb{z}}_{1:d};x,t)$. We define a likelihood loss function for the generator as:

\begin{equation}
\label{eqn:loss_nll}
    \begin{aligned}
    \mathcal{L}_{\text{NLL}}(\theta) &= -\sum_{i=1}^{N_{\omega}}  \sum_{(x_o,t_o) \in O} \log p_{\theta}(\pmb{u}|x_o,t_o,\omega^{(i)}) \\
    & = -\sum_{i=1}^{N_{\omega}}  \sum_{(x_o,t_o) \in O} \log p_{\tilde{\pmb{z}}}(\tilde{\pmb{z}}|x_o,t_o) + \sum_{l=0}^{L-1} \log \left|\operatorname{det}\left(\frac{\partial f_{l}^{-1}(\pmb{h}^{(l)})}{\partial \pmb{h}^{(l)}}\right)\right|^{-1}
    ,
    \end{aligned}
\end{equation}
which is the summation of negative log-likelihood (NLL) over all observed coordinates and random events.

Apart from the likelihood loss $\mathcal{L}_{\text{NLL}}$, the flow generator $f_{\theta}$ can also be trained with the discriminator $D_{\phi}$ through adversarial training. The adversarial loss for the generator is depicted as:
\begin{equation} \label{eqn:loss_adv}
    \mathcal{L}_{\text{Adv}}(\theta) = -\frac{1}{N_{\omega}} \sum_{i=1}^{N_{\omega}} D_{\phi}(\hat{M}^{(i)}),
\end{equation}
which uses the Wasserstein objective defined in Eq.~\ref{eqn:wgan}; $\hat{M}$ is the prediction matrix and $N_{\omega}$ is the total number of sampling.

Using the adversarial loss $\mathcal{L}_{\text{Adv}}$ alone is prone to mode collapse. We demonstrate below how the likelihood loss $\mathcal{L}_{\text{NLL}}$ can mitigate the mode collapse by one example. Suppose the data is generated from a mixture model of two Gaussian distributions $\mathcal{N}(-1,1)$ and $\mathcal{N}(1,1)$. By adversarial training, the generator may end up only generating one mode, say $\mathcal{N}(-1,1)$. In this case, the discriminator cannot distinguish between the samples and the ground truth. If the MLE is used, the likelihood of the missing mode is very low, resulting in a high overall NLL. Thus, the likelihood loss $\mathcal{L}_{\text{NLL}}$ can guide the generator to leave the current local optimum.

\subsection{Physics Regularization}

The conditional flow model is further regularized by the physics-informed computational graph, which encodes physics prior knowledge like partial differential equations (PDE). 


Suppose data follows laws that can be depicted as stochastic PDEs below:
\begin{equation}
\label{equ:pde}
\begin{aligned}
    \pmb{u}_t(x,t; {\omega}) + \mathcal{N}_x[\pmb{u}(x,t;\omega);{\lambda}(\omega)] &= \pmb{0}, 
    \quad
    (x,t) \in \mathcal{X} \times \mathcal{T},
    \omega \in \Omega,
    \\
    \mathcal{B}[\pmb{u}(x,t;\omega)] &= \pmb{0},
    \quad
    (x,t) \in \partial \mathcal{X} \times \mathcal{T},
    \\
    \mathcal{I}[\pmb{u}(x,0; \omega)] &= \pmb{0},
    \quad x \in \mathcal{X},
\end{aligned}
\end{equation}
where, $\pmb{u}_t$ is its partial derivative of $\pmb{u}$ with regard to $t$; $\partial \mathcal{X}$ is the boundary of the space domain $\mathcal{X}$; $\mathcal{N}_x$ is a non-linear differential operator; $\mathcal{B}$ is a boundary condition operator; $\mathcal{I}$ is an initial condition operator; ${\lambda}$ is the parameters of the PDEs. $\omega$ is a random event sampled from the probability space $\Omega$, which represents uncertainties residing in the PDE parameters or the boundary and initial conditions.

By encoding physics information, the physics-informed flow generator has an additional learning objective on the unobserved region $C$, which encourages the prediction of the generator to follow the physics defined by the PDE. The physics loss function is defined as:

\begin{equation}\label{eqn:phys_loss}
\mathcal{L}_{\text{Phy}}(\theta, {\lambda}) = \mathbb{E}_{q(x_c, t_c)}  \left| \mathbb{E}_{p_{\pmb{z}}(\pmb{z})} \left [  (\hat{\pmb{u}}_c)_t + \mathcal{N}_x[\hat{\pmb{u}}_c; {{\lambda}}] \right] \right|^2, 
\end{equation}
where $\hat{\pmb{u}}_c = f_{\theta}(x_c, t_c, \pmb{z})$ is the prediction of the generator on the unobserved region. This physics loss function serves as a regularization term for the generator. If the flow generator $f_{\theta}$ is well trained, the
physics loss needs to be as close to zero as possible.


\subsection{Training of TrafficFlowGAN}
\label{sec:training}

The loss function of the flow model is a weighted sum of the likelihood loss, adversarial loss, and physics loss:
\begin{equation} \label{eqn:loss_f}
 \mathcal{L}_{f}(\theta) = \alpha \mathcal{L}_{\text{NLL}}(\theta) + \beta \mathcal{L}_{\text{Adv}}(\theta) + \gamma  \mathcal{L}_{\text{Phy}}(\theta, {\lambda}),
\end{equation}
where $\alpha, \beta, \gamma \in (0,1]$ are hyperparameters that determine the contribution of each loss component. With the generator loss $\mathcal{L}_{f}(\theta)$, the discriminator loss $\mathcal{L}_{D}(\phi)$ defined in Eq.~\ref{eqn:loss_d}, and physics loss $\mathcal{L}_{\text{Phy}}(\theta, {\lambda})$ defined in Eq.~\ref{eqn:phys_loss}, we are ready to introduce the training algorithm as shown in Algorithm~\ref{alg}.  \vspace{-4mm}

\begin{algorithm}[h]
\caption{TrafficFlowGAN Training Algorithm.}
\label{alg}
\textbf{Initialization}:\\ 
Initialized physics parameters ${\lambda}^0$;
Initialized networks parameters $\theta^0$, $\phi^0$;
Training iterations $Iter$;
Batch size $m$;
Learning rate $lr$;
Weights of loss functions $\alpha$, $\beta$, and $\gamma$.\\
\textbf{Input}: The observation data $\{(x_o^{(i)},t_o^{(i)},\pmb{u}_o^{(i)})\}_{i=1}^{N_o}$ and unobserved coordinates $\{x_c^{(j)},t_c^{(j)}\}_{j=1}^{N_c}$.
\begin{algorithmic}[1] 
\label{alg:train}
\FOR{$k \in \{0,...,Iter\}$}
\STATE Sample batches $\{(x_o^{(i)},t_o^{(i)},\pmb{u}_o^{(i)})\}_{i=1}^{m}$ and $\{x_c^{(j)},t_c^{(j)}\}_{j=1}^{m}$ from the observation data and unobserved coordinates, respectively \\
{\mycommfont{// update the discriminator}}\\
\STATE Calculate ${\cal L}_{D}$ by Eq.~\ref{eqn:loss_d} \\
\STATE $\phi^{k+1} \leftarrow \phi^{k} -lr \cdot \text{Adam}(\phi^{k},\nabla_{\phi} {\cal L}_D)$ \\
{\mycommfont{// update the generator}}\\
\STATE Calculate ${\cal L}_{\text{NLL}}$ by Eq.~\ref{eqn:loss_nll}, ${\cal L}_{\text{Adv}}$ by Eq.~\ref{eqn:loss_adv}, and ${\cal L}_{\text{Phy}}$ by Eq.~\ref{eqn:phys_loss}\\
\STATE Calculate ${\cal L}_{f}$ by Eq.~\ref{eqn:loss_f} \\
\STATE $\theta^{k+1} \leftarrow \theta^{k}-lr\cdot\text{Adam}(\theta^{k},\nabla_{\theta} {\cal L}_{f})$ \\
{\mycommfont{// update the physics}}\\
\STATE $\lambda^{k+1} \leftarrow \lambda^{k}-lr\cdot\text{Adam}(\lambda^{k},\nabla_{\lambda} {\cal L}_{\text{Phy}})$ \\
\ENDFOR
\end{algorithmic}
\end{algorithm}

\section{Numerical Experiment: Learning Solutions of A Known Second-order PDE} \label{sec:experiment}

In this experiment, we apply TrafficFlowGAN to learn solutions of a known PDE and also to learn the relations of the PDE's parameters. 

\subsection{Numerical Data}
\begin{figure}%
    \centering
    \subfloat[\centering ]{{\includegraphics[height=0.25\columnwidth]{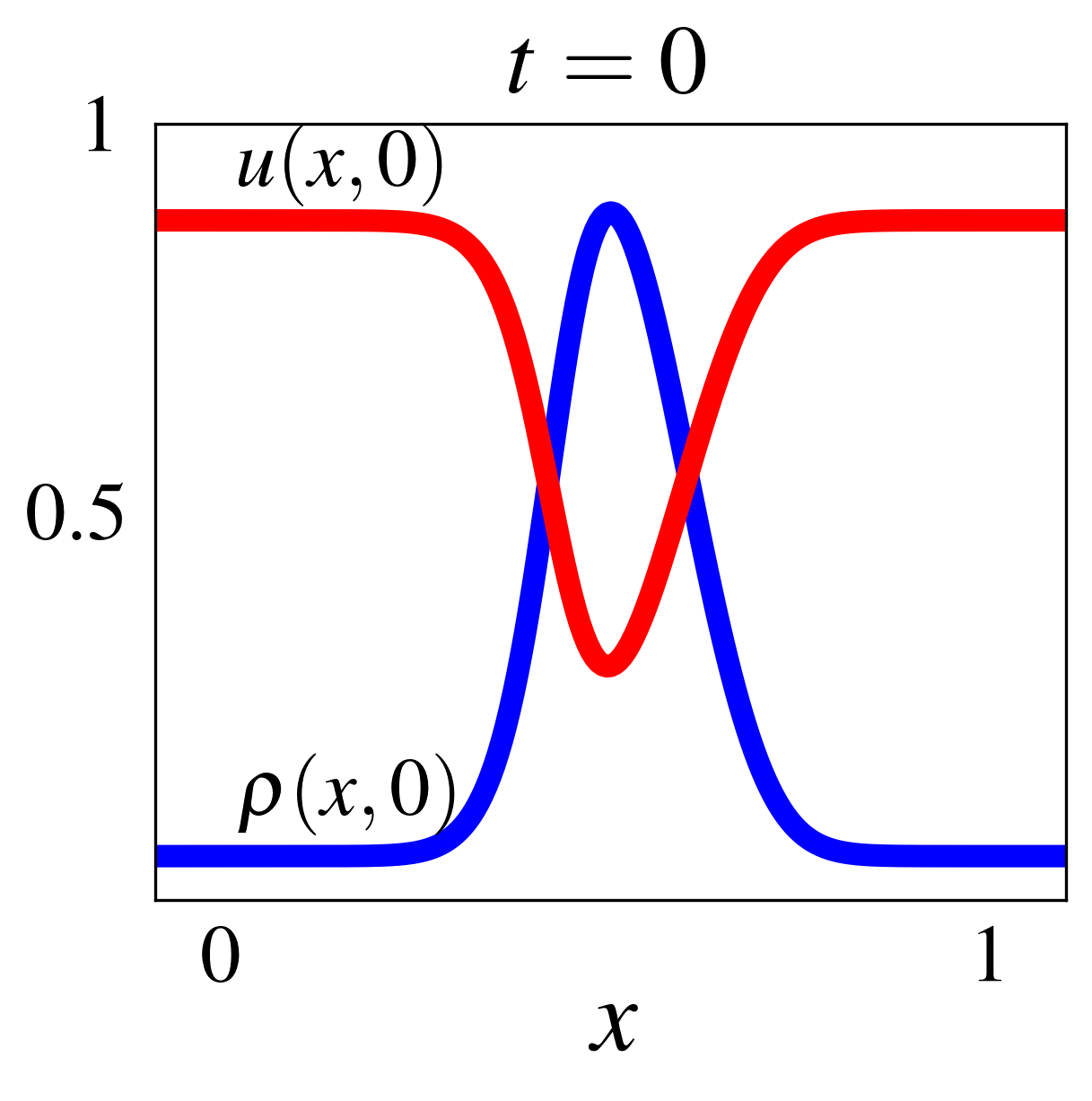} }}\hspace{10mm}
    \subfloat[\centering ]{{\includegraphics[height=0.25\columnwidth]{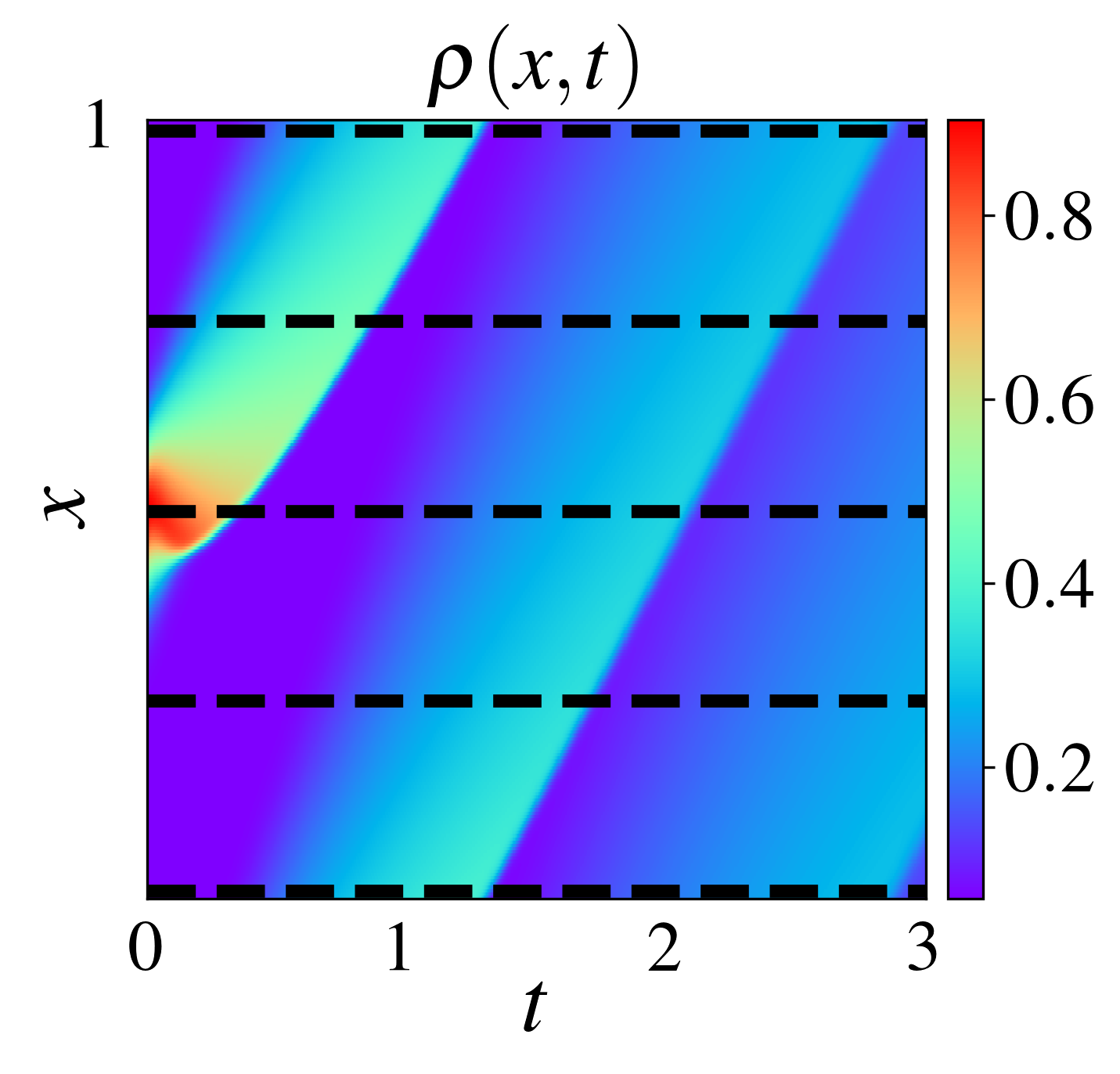} }}\hspace{10mm}
     \subfloat[\centering ]{{\includegraphics[height=0.25\columnwidth]{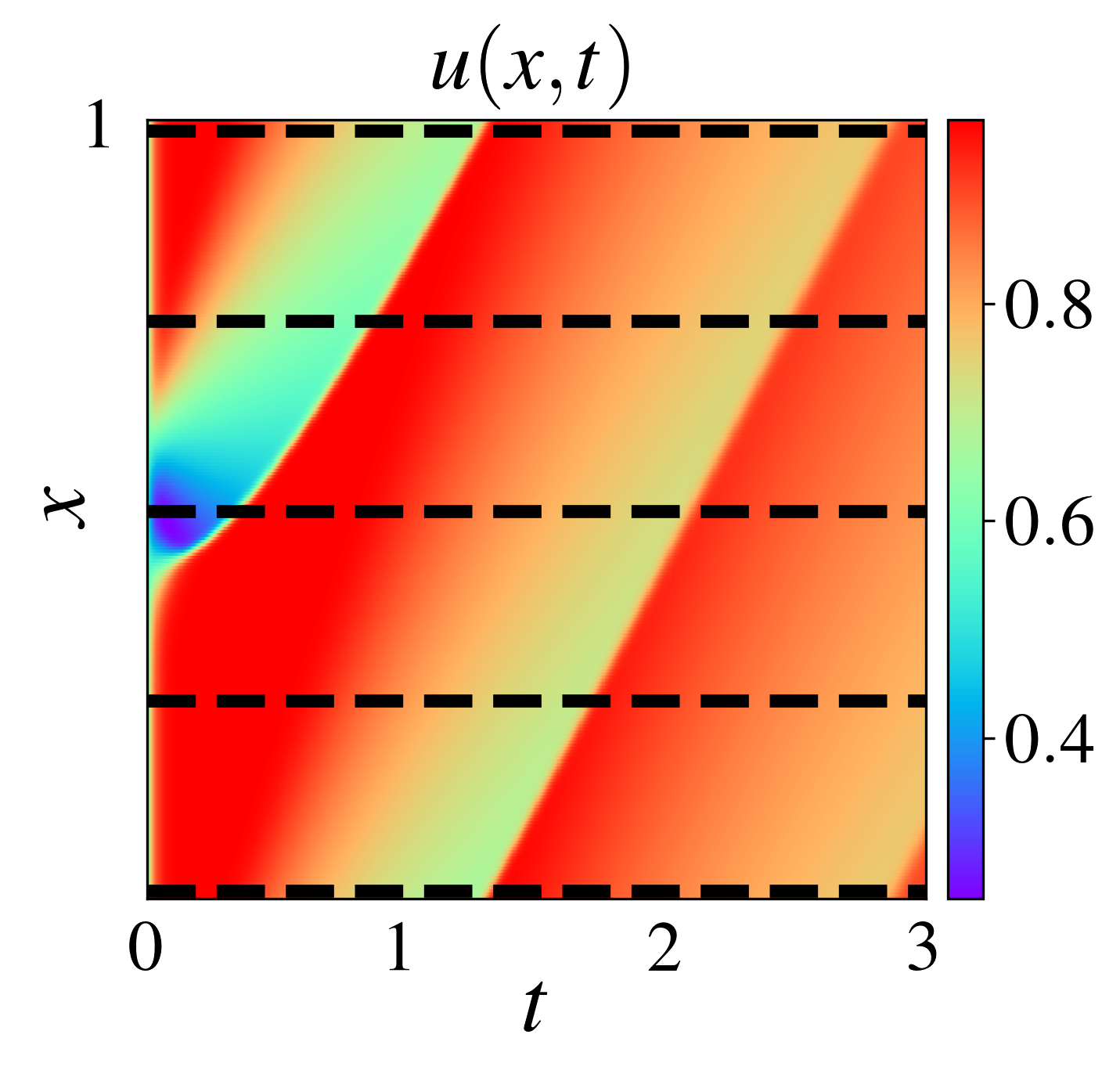} }}\hspace{10mm}
    \caption{ Numerical data generator from the ARZ model. (a) is the bell-shaped initial $\rho$ and $u$ over $x\in [0,1]$; (b) and (c) are numerical solutions for $\rho$ and $u$, respectively.}%
    \label{fig:ch5-initial-r3}%
\end{figure}

The numerical data is generated from the ARZ model \cite{Aw-Rascle-2000}, which is a second-order PDE that is used to describe the traffic dynamics. It is depicted as

\begin{equation}\label{eqn:arz}
\begin{aligned}
\begin{cases}
    \rho_t + (\rho u)_x = 0, \\
    (u+h(\rho))_t + u(u+h(\rho))_x = (U_{eq}(\rho)-u)/ \tau, 
  \end{cases} 
\end{aligned}
\end{equation}
where, 

\begin{equation}
    h(\rho) = U_{eq}(0) - U_{eq}(\rho)
\end{equation}
is the hesitation function and 
\begin{equation}
\label{eqn:arz_ueq}
U_{eq}(\rho) = u_{max}(1 - \rho / \rho_{max})
\end{equation}
is the equilibrium traffic velocity;
traffic density $\rho$ and traffic velocity $u$ are the state variables, i.e. $\pmb{u}=(\rho, u)$; $\tau$ is the relaxation parameter; $\rho_{max}$ and $u_{max}$ are the maximum traffic density and the maximum traffic velocity, respectively. In this experiment, we study a ``ring road'' in $t \in [0,3]$ and $x \in [0,1]$ with a boundary condition $\pmb{u}(0,t)=\pmb{u}(1,t)$. We set the parameters as $\rho_{max}=1.13$, $u_{max}=1.02$, and $\tau=0.02$. We set the initial conditions of $\rho$ and $u$ as bell-shaped functions shown in Fig. \ref{fig:ch5-initial-r3}(a). The x-axis is the space domain, and the y-axis is the initial value of $\rho$ (blue line) and $u$ (red line). We solve Eq.~\ref{eqn:arz} using the Lax-Friedrichs scheme on a spatio-temporal grid of sizes $240\times960$, and the solutions $\rho(x,t)$ and $u(x,t)$ are shown in Fig.~\ref{fig:ch5-initial-r3}(b) and Fig.~\ref{fig:ch5-initial-r3}(c), respectively. The dashed black lines indicate 4 locations where data is observed (the top and bottom dashed black lines indicate the same position as it is a ring road). We then add a white noise $\epsilon \sim \mathcal{N}(0,0.02)$ to the solution to represent the uncertainty.   
\vspace{-5mm}

\vspace{-10mm}
\subsection{Physics-based Computational Graph}
\begin{figure}%
    \centering
    \subfloat[\centering ]{{\includegraphics[width=0.3\columnwidth]{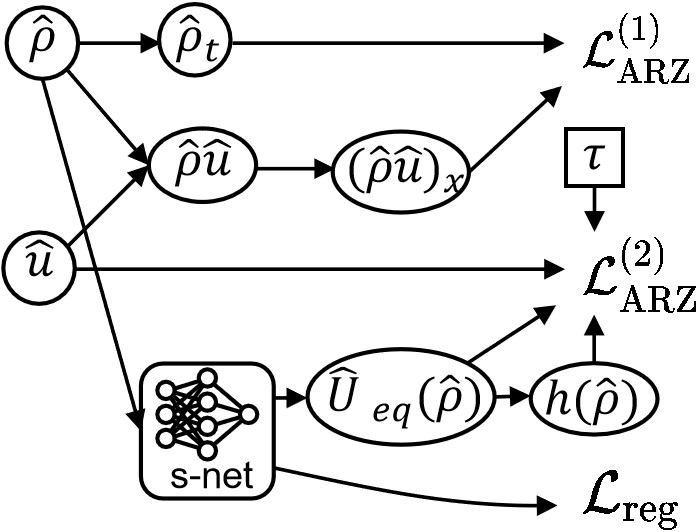} }}\hspace{20mm}
    \subfloat[\centering ]{{\includegraphics[width=0.3\columnwidth]{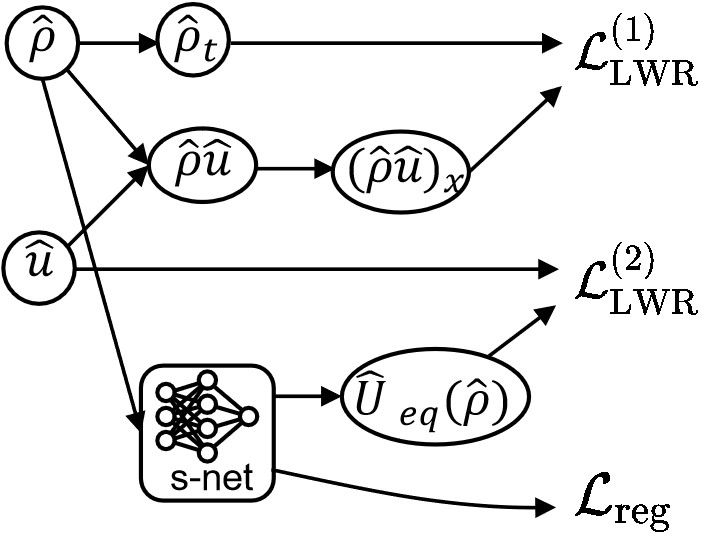} }}%
    \caption{Physics with surrogate models. (a) is the physics for ARZ and (b) is the physics for LWR, which will be introduced in Section~\ref{sec:phys_lwr}. }%
    \label{fig:physics_with_FD}%
\end{figure}

Fig.~\ref{fig:physics_with_FD}(a) illustrates the physics-based computational graph assosciated with the ARZ. We assume that the exact form of $U_{eq}$ is unknown, and we use a \emph{surrogate network} (s-net) to learn an approximate $\hat{U}_{eq}$. The corresponding physics loss for each line of Eq.~\ref{eqn:arz} is as below:
\begin{equation}\label{eqn:phyloss_arz}
\begin{aligned}
\begin{cases}
    \mathcal{L}_{\text{ARZ}}^{(1)} = \left| \mathbb{E}_{\pmb{z}} \left[ \hat{\rho}_t + (\hat{\rho}\hat{u})_x\right]  \right|^2 \\
     L_{\text{ARZ}}^{(2)} = \left| \mathbb{E}_{\pmb{z}} \left[  (\hat{u}+h(\hat{\rho}))_t + \hat{u}(\hat{u}+h(\hat{\rho}))_x - (\hat{U}_{eq}(\hat{\rho})-\hat{u}) / \tau  \right]  \right|^2.
  \end{cases}
\end{aligned}
\end{equation}
In implementation, derivatives with regard to $x$ and $t$ can be easily calculated by the Pytorch module {\ttfamily \footnotesize torch.autograd}.

Additionally, we add a \emph{shape constraint} to regularize the s-net to be monotonically decreasing for the ARZ physics loss, given the domain knowledge that the equilibrium speed $U_{eq}$ decreases as the density $\rho$ increases. This shape constraint is depicted as follows:
\begin{equation} \label{eqn:phyloss_ref}
    \mathcal{L}_{\text{reg}}=\int_{a}^{b} \max \left(0, \frac{\partial \hat{U}_{eq}(\rho)}{\partial \rho}\right) d \rho,
\end{equation}
where hyperparameters $a$ and $b$ determine the interval where the shape constraint takes effect, e.g. $a=0$ and $b=1.13$ in this ring road experiment. Summarizing the loss terms defined in Eq.~\ref{eqn:phyloss_arz} and Eq.~\ref{eqn:phyloss_ref}, the final physics loss can thus be written as:
\begin{equation}
    \mathcal{L}_{\text{Phy}} = \eta \mathcal{L}_{\text{ARZ}}^{(1)} + (1- \eta) \mathcal{L}_{\text{ARZ}}^{(2)} + \xi \mathcal{L}_{\text{reg}},
\end{equation}
where $\eta \in (0,1]$ and $\xi \in (0,\infty)$ are hyperparameters that control the weights.

\subsection{Experiment Setting}

Experiments are conducted on a Google cloud workstation with 8 Intel Xeon E5-2686 v4 processors and an NVIDIA V100 Tensor Core GPU with 16 GB memory in Ubuntu 18.04.3. 
The learning rate for the Adam optimizer is $0.0005$, and other configurations are kept as default. The configuration of the discriminator is different for different loop detectors, which are detailed in the supplementary materials.

\subsection{Results}
\begin{figure}%
    \centering
    \subfloat[\centering ]{{\includegraphics[height=0.25\columnwidth]{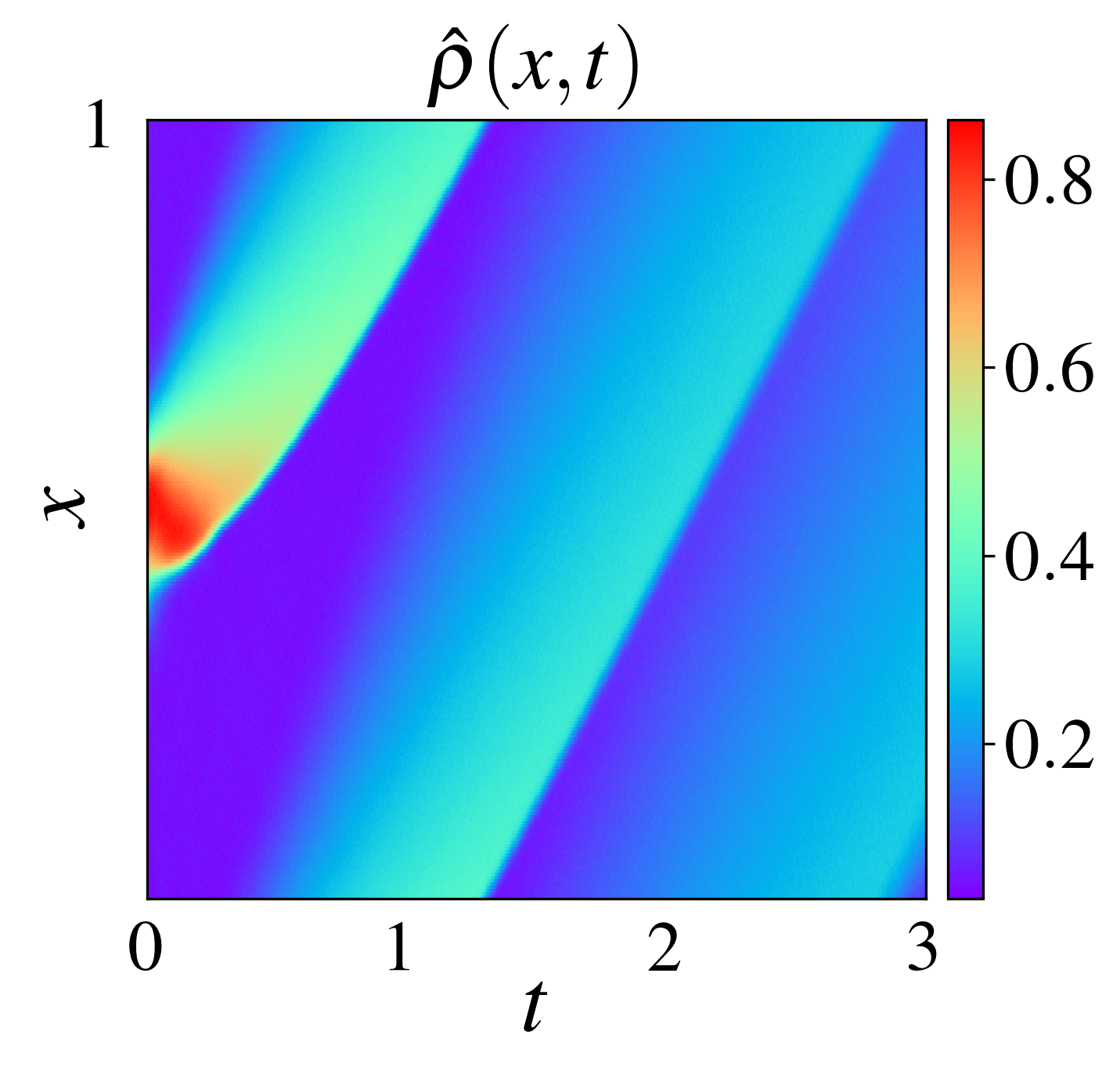} }}%
    \subfloat[\centering ]{{\includegraphics[height=0.25\columnwidth]{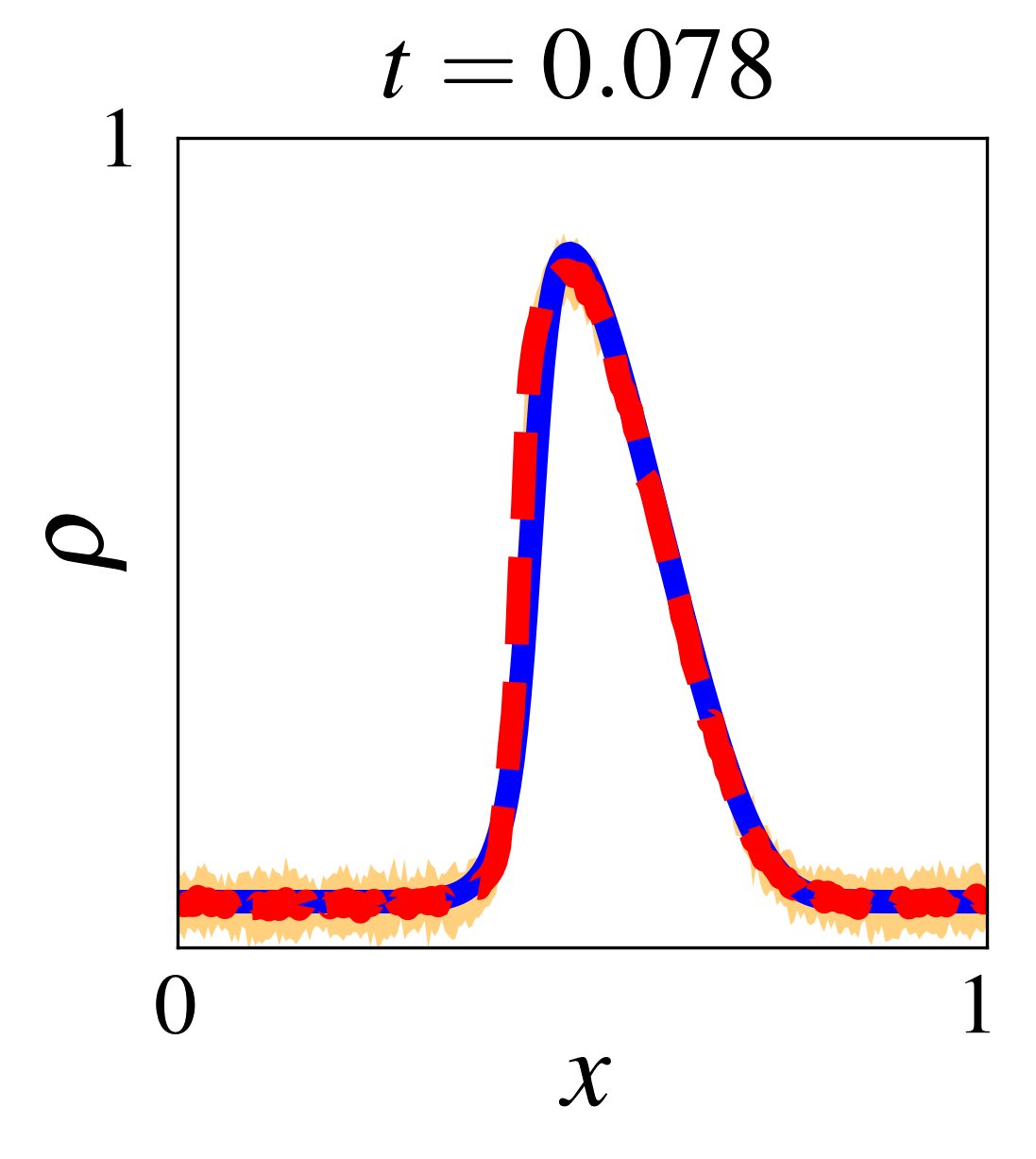} }}%
     \subfloat[\centering ]{{\includegraphics[height=0.25\columnwidth]{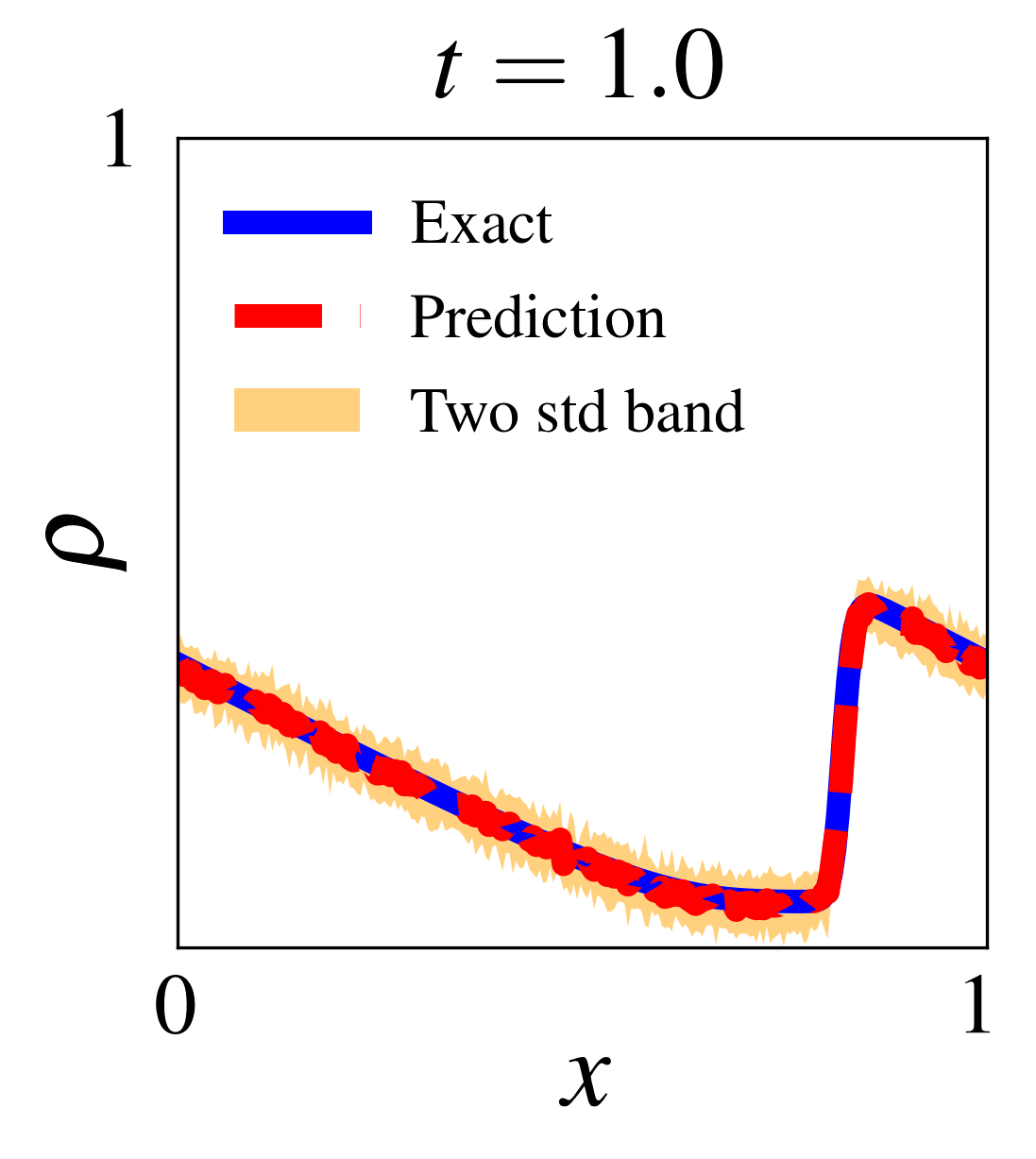} }}%
     \subfloat[\centering ]{{\includegraphics[height=0.232\columnwidth]{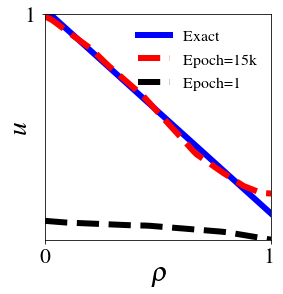} }}%
    \caption{ Results of the TrafficFlowGAN on the numerical data generated by ARZ.}%
    \label{fig:result_numerical}%
\end{figure}

The results of the TrafficFlowGAN are shown in Fig.~\ref{fig:result_numerical}. Fig.~\ref{fig:result_numerical}(a) is the prediction of the traffic density. It demonstrates that the TrafficFlowGAN can reconstruct the traffic density with observations from 4 sensors. Two prediction snapshots at $t=0.078$ and $t=1.0$ shown in Fig.~\ref{fig:result_numerical}(b) and Fig.~\ref{fig:result_numerical}(c), respectively. The blue line stands for the mean of the ground truth; the dashed red line represents the mean of the prediction, and the yellow band is the prediction interval. Fig.~\ref{fig:result_numerical}(d) illustrates the relation between traffic density and traffic velocity learned by the s-net, i.e. $\hat{U}_{eq}(\hat{\rho})$. The solid blue line is the ground-truth relation $U_{eq}(\rho)$ that is defined in Eq.~\ref{eqn:arz_ueq}. The dashed black line and the dashed red line are the $\hat{U}_{eq}(\hat{\rho})$ at the 1st and the 15000th epochs, respectively. We can see that s-net manages to recover the underlying traffic density-velocity relation. The reason for the relatively poor performance for $\rho > 0.9$ is that the numerical data does not contain $\rho$ that is bigger than 0.9, as indicated by the colorbar of Fig.~\ref{fig:ch5-initial-r3}(b).

\vspace{-2mm}

\section{Case Study: Traffic State Estimation} \label{sec:case_study}
Traffic state estimation (TSE) is an important traffic engineering problem that aims to infer traffic state variables represented by traffic density and velocity along a road segment from partial observations. 
In a nutshell, the goal of TSE is to learn a mapping from a spatio-temporal domain to traffic states, i.e. $f:(x,t) \rightarrow (\rho, u)$, using partial observations from fixed sensors (e.g. loop detectors).

\subsection{Dataset}
\begin{figure}%
    \centering
    \subfloat[\centering ]{{\includegraphics[width=0.49\columnwidth]{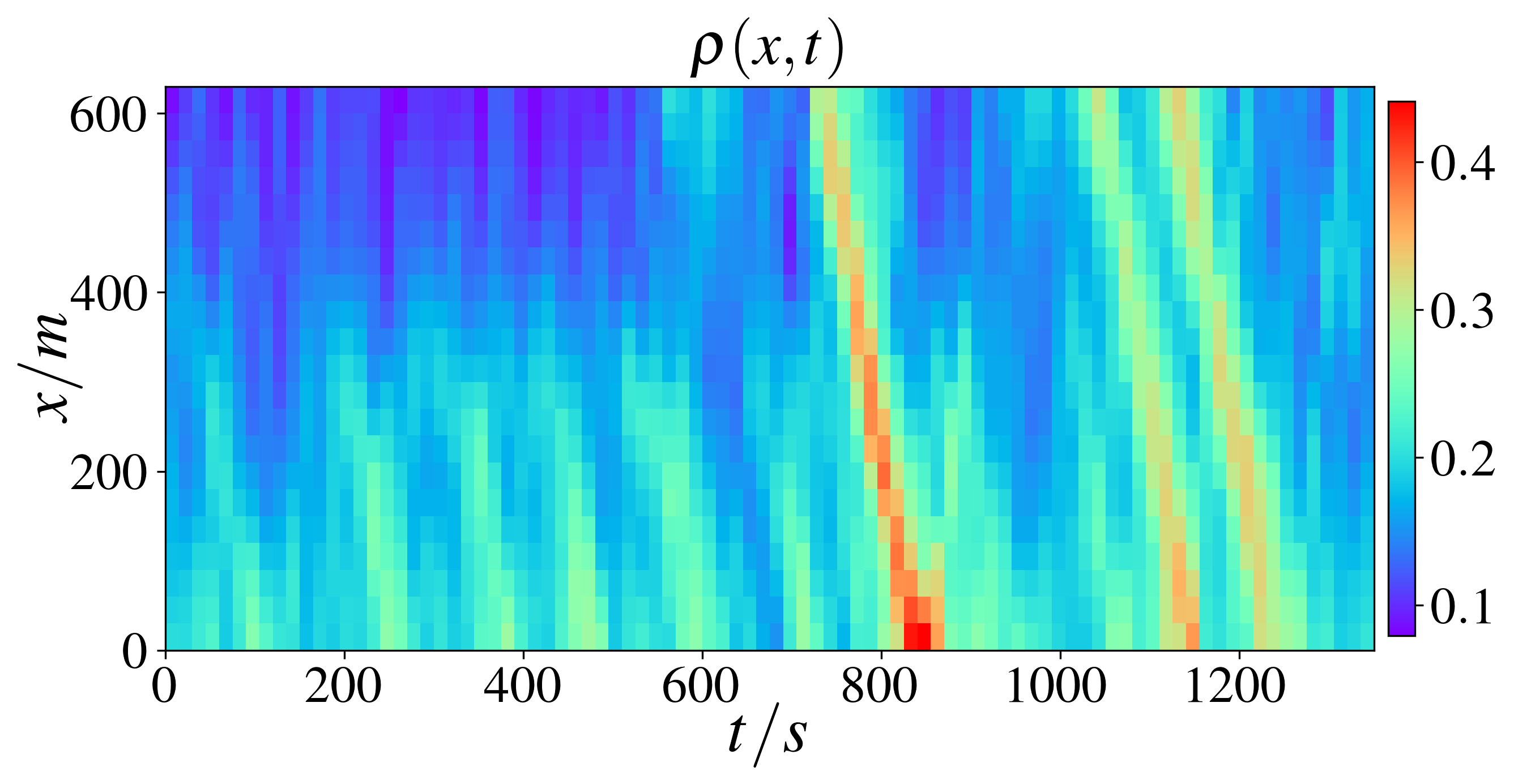} }} 
    \subfloat[\centering ]{{\includegraphics[width=0.49\columnwidth]{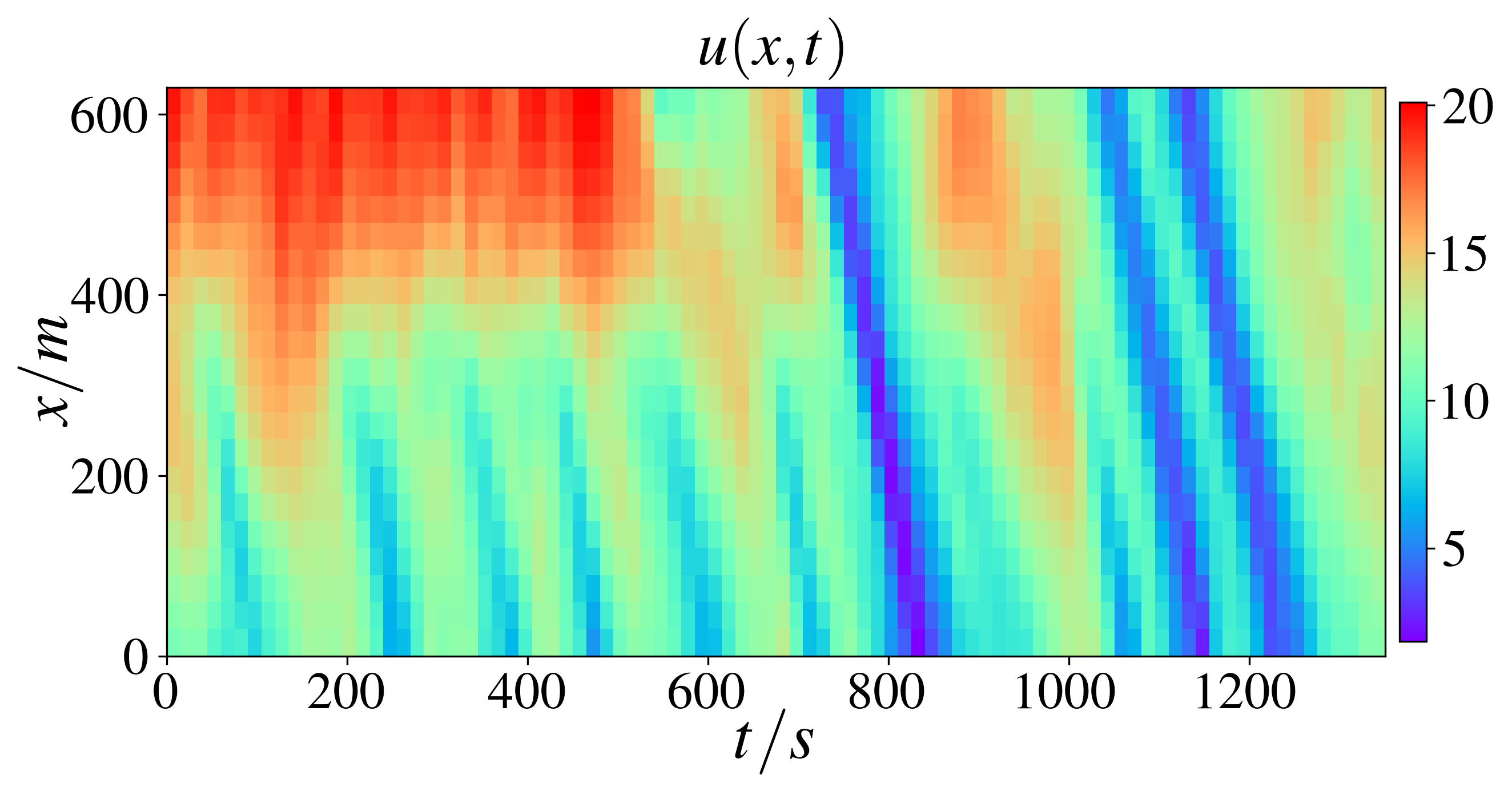} }}%
    \caption{ NGSIM dataset, where (a) is the traffic density and (b) is the traffic velocity.}%
    \label{fig:ngsim_data}%
\end{figure}

The Next Generation SIMulation (NGSIM)\footnote{www.fhwa.dot.gov/publications/research/operations/07030/index.cfm} is a real-world dataset that collects vehicle trajectory every 0.1 seconds. We focus on a 15-minutes data fragments collected on highway US 101. The traffic density and velocity are shown in Fig.~\ref{fig:ngsim_data}. 
\vspace{-14mm}

\vspace{-5mm}
\subsection{Physics-based Computational Graph} \label{sec:phys_lwr}

As the underlying physics for the real-world scenario is unknown, in addition to the ARZ, we also adopt the LWR model as our physics. LWR is depicted as below:
\begin{equation}\label{eqn:lwr}
\begin{aligned}
\begin{cases}
    \rho_t + (\rho u)_x =  0, \\
    u= U_{eq}(\rho) \triangleq u_{max}(1 - \rho / \rho_{max}),
  \end{cases}
\end{aligned}
\end{equation}
which shares the same physics parameters, i.e. $\rho_{max}$ and $u_{max}$, as the ARZ. The physics computational graph associated with LWR is illustrated in Fig.~\ref{fig:physics_with_FD}(b). The corresponding physics losses are as below:
\begin{equation}\label{eqn:phyloss_lwr}
\begin{aligned}
\begin{cases}
    \mathcal{L}_{\text{LWR}}^{(1)} = \left| \mathbb{E}_{\pmb{z}} \left[ \hat{\rho}_t + (\hat{\rho}\hat{u})_x\right]  \right|^2 \\
     L_{\text{LWR}}^{(2)} = \left| \mathbb{E}_{\pmb{z}} \left[  (\hat{U}_{eq}(\hat{\rho})-\hat{u})   \right]  \right|^2
  \end{cases}
\end{aligned}.
\end{equation}




\subsection{Baselines and Metrics}

We adopt the following baselines for comparison: the pure flow model, the physics-informed flow with ARZ as the physics (PhysFlow-ARZ), the physics-informed flow with LWR as the physics (PhysFlow-LWR), FLowGAN, and the ARZ-based extended Kalman filter (EKF) \cite{seo2017traffic}. EKF applies a non-
linear version of the Kalman filter and is widely used
in nonlinear systems like the TSE.

We use the $\mathbb{L}^2$ relative percentage error ({RE}) to measure the difference between the mean of the prediction and that of the ground truth. The reason for choosing this metric is to mitigate the influence of the scale of the ground truth.  In addition, the reverse Kullback-Leibler ({KL}) divergence is used to measure the difference between the prediction distribution and the sample distribution.

\subsection{Results}

\begin{figure}[h!]
  \centering
  \includegraphics[width=\linewidth]{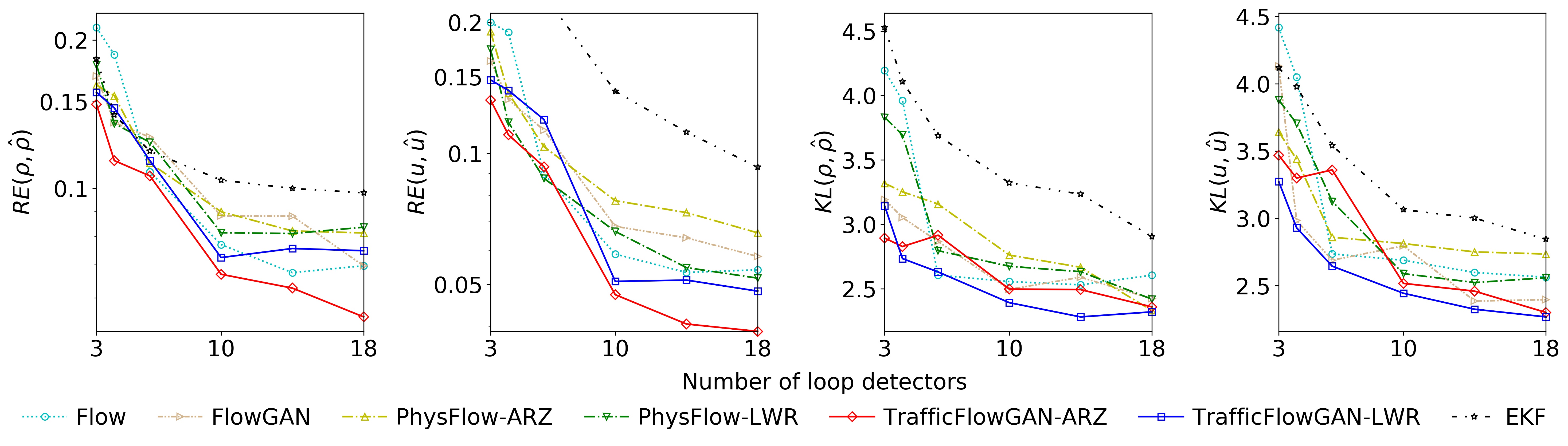}
  \caption{The REs and KL of our proposed TraficFlowGAN and the baselines.}
  \label{fig:ngsim_result_REKL}
\end{figure}

Fig.~\ref{fig:ngsim_result_REKL} shows the REs (left two) and KL divergences (right two) of traffic density $\rho$ and velocity $u$ of TrafficFlowGAN and the baselines. The x-axis is the number of loop detectors. Different scatter types and colors are used to distinguish with different models. From this figure, we can see that TrafficFlowGAN-ARZ outperforms others across nearly all numbers of loop detectors for REs, and TrafficFlowGAN-LWR achieves the best performance for KL divergences. We also record the training time of each model when the number of loops is 10. The Flow-based models, including Flow and PhysFlow, cost 0.11 second per epoch. This means that the extra computational time from calculating the physics loss is negligible. The training time of the FlowGAN-based models, including TrafficFlowGAN and FlowGAN, is 0.31 second per epoch. 

\begin{figure}[h!]%
    \centering
    \subfloat[\centering ]{{\includegraphics[height=0.25\columnwidth]{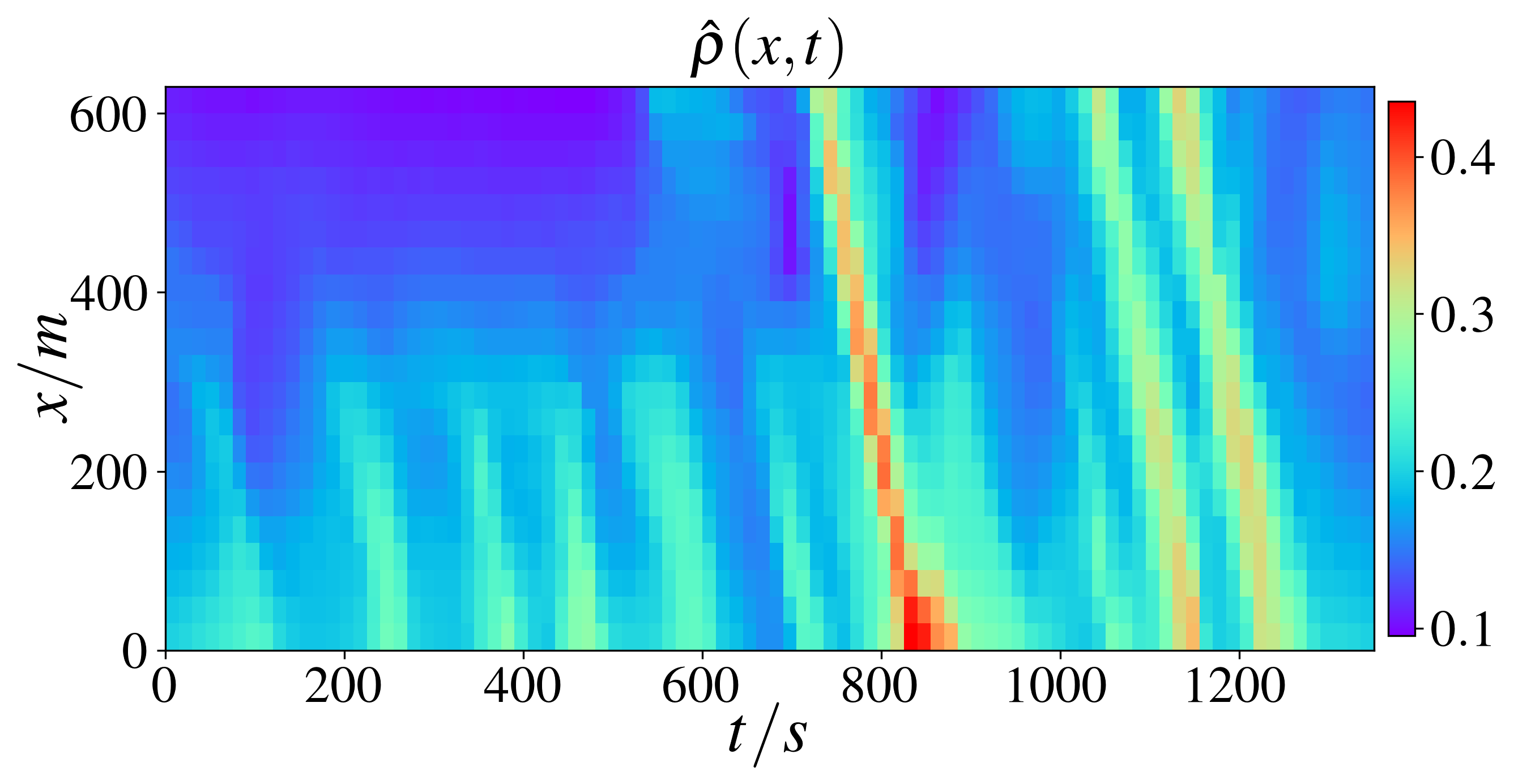} }}%
    \subfloat[\centering ]{{\includegraphics[height=0.25\columnwidth]{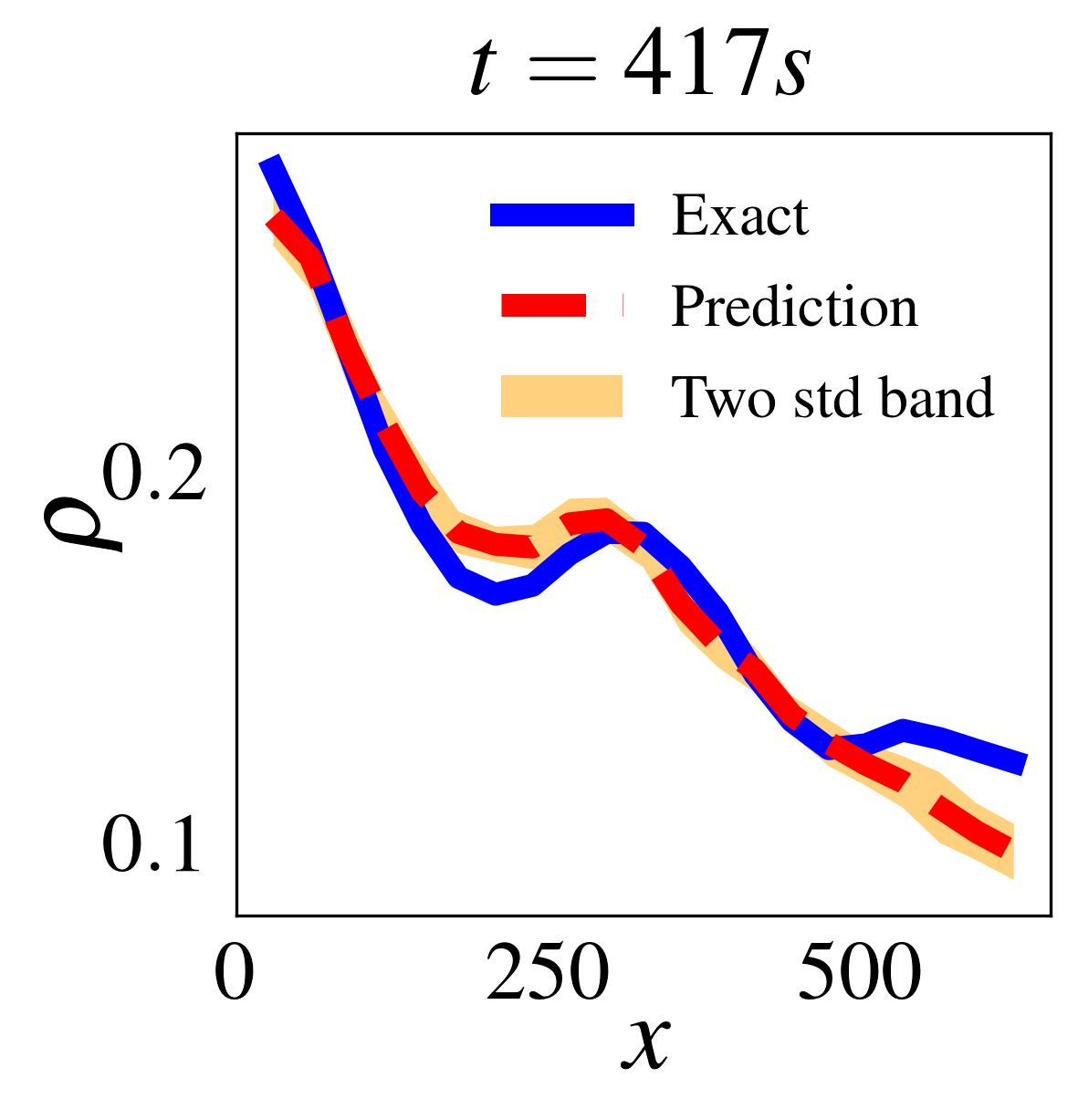} }}%
     \subfloat[\centering ]{{\includegraphics[height=0.25\columnwidth]{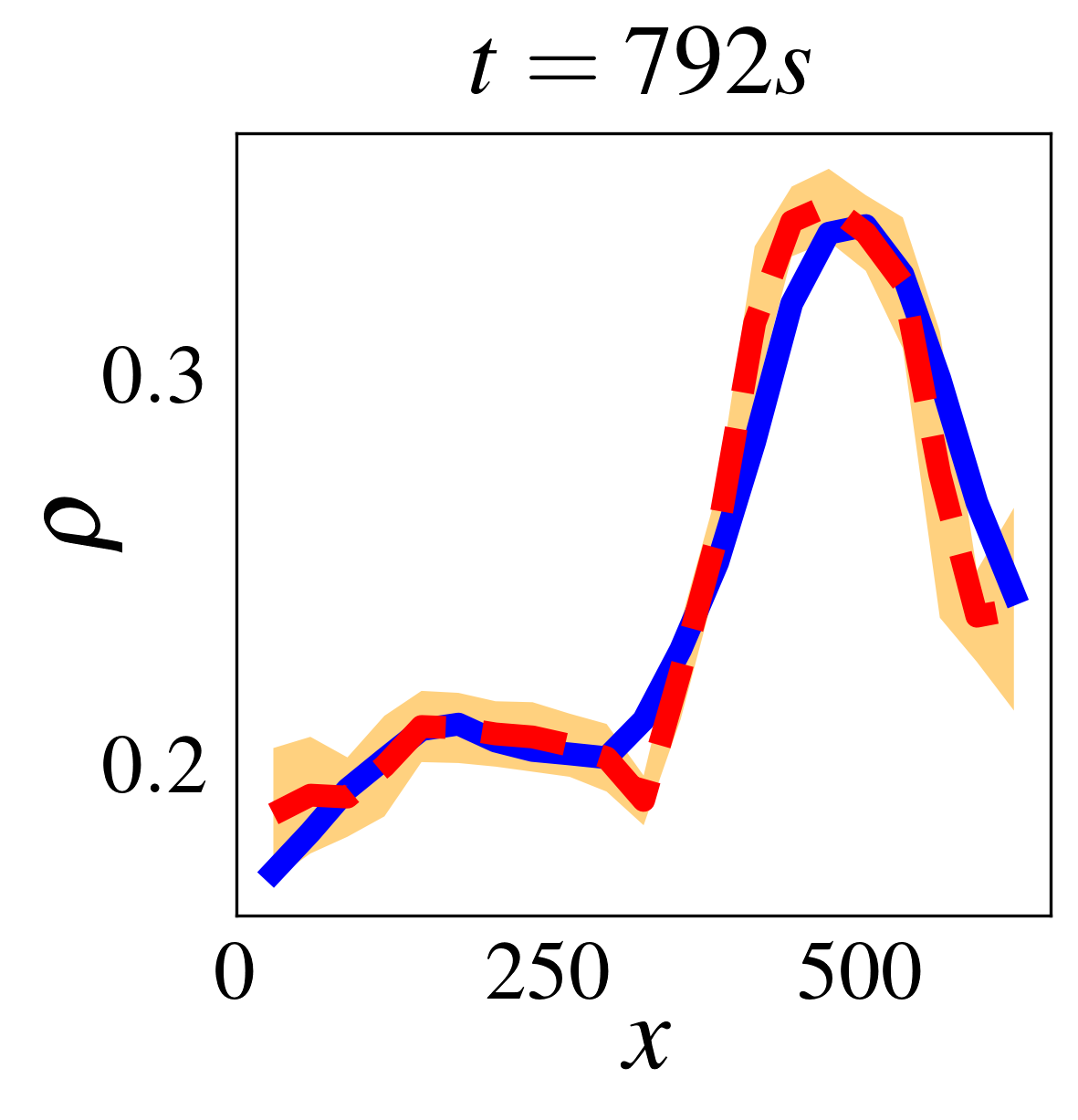} }} \\ \vspace{-4mm}
     \subfloat[\centering ]{{\includegraphics[height=0.25\columnwidth]{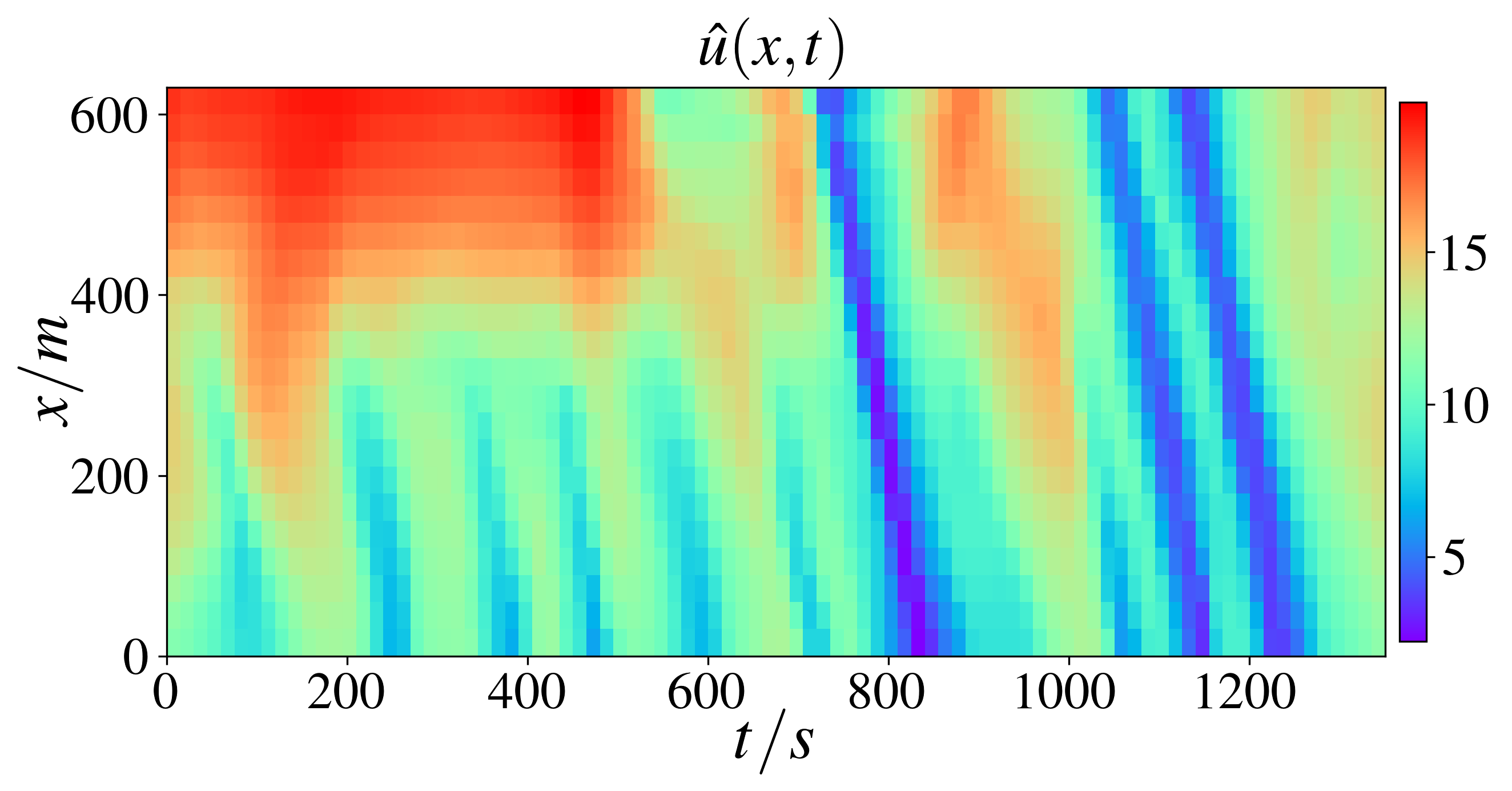} }} 
    \subfloat[\centering ]{{\includegraphics[height=0.24\columnwidth]{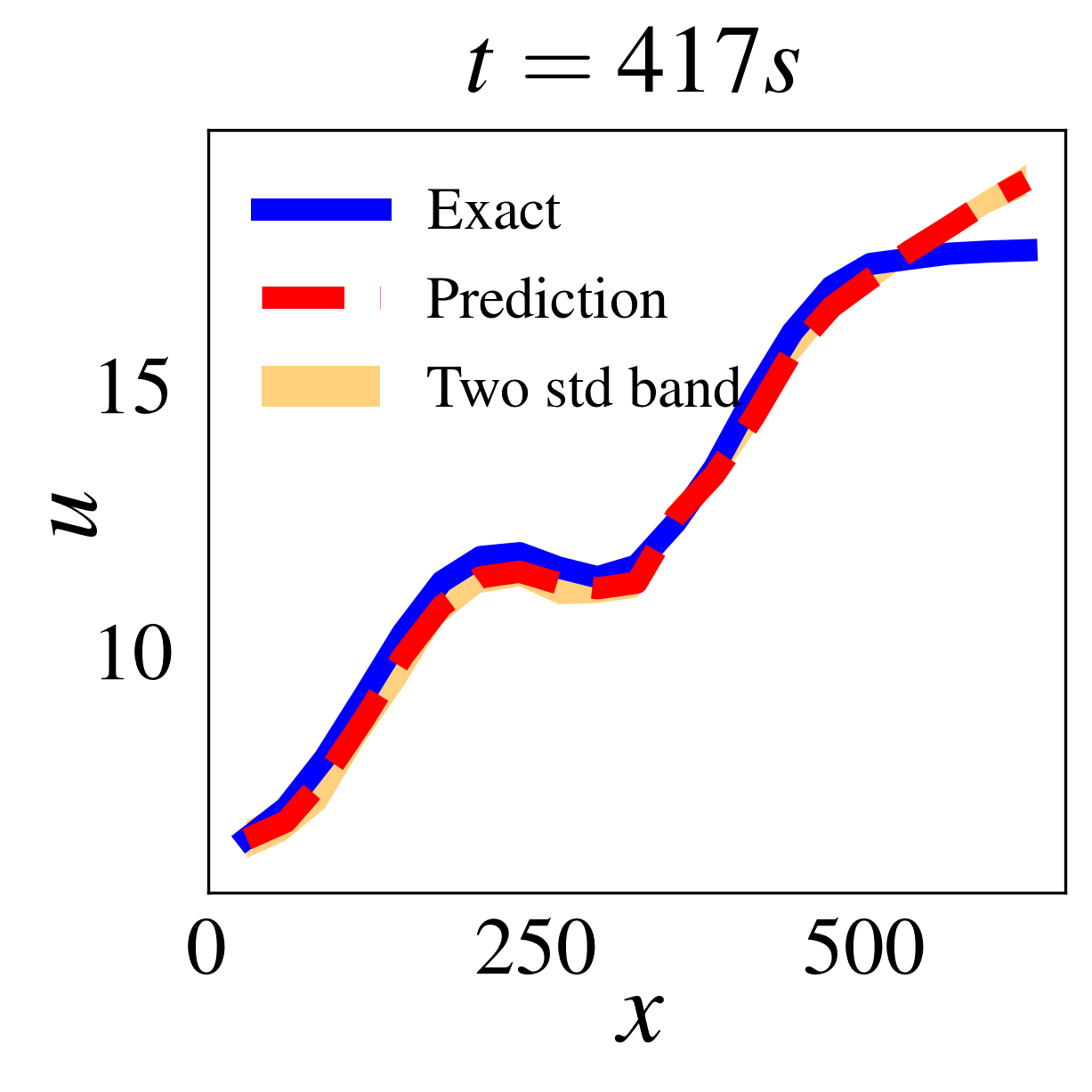} }}%
     \subfloat[\centering ]{{\includegraphics[height=0.24\columnwidth]{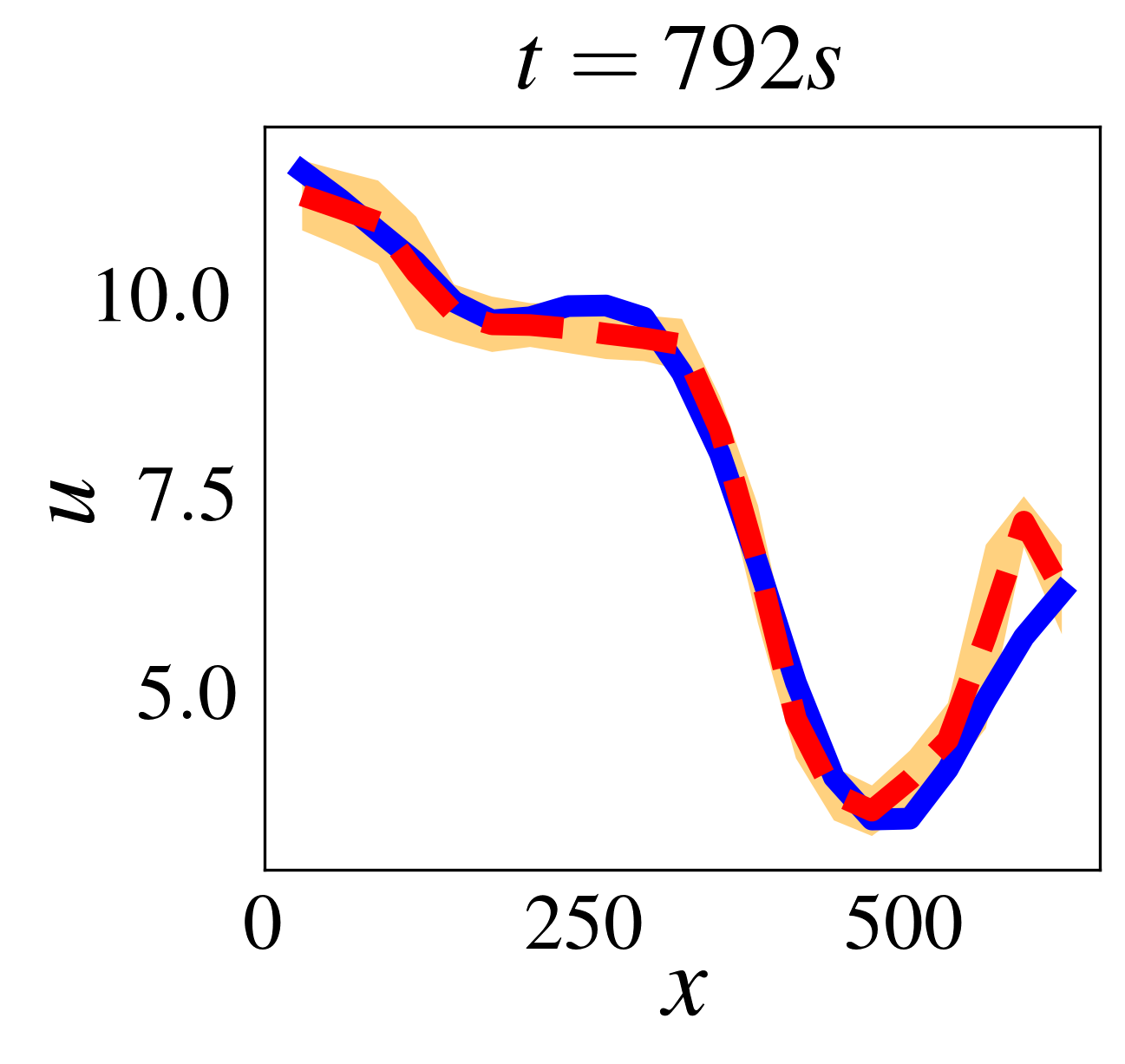} }}%
     
    \caption{ Predictions of the traffic density (top row) and the traffic velocity (bottom row) of the TrafficFlowGAN.}%
    \label{fig:ngsim_result_prediction}%
\end{figure}

 Fig.~\ref{fig:ngsim_result_prediction} show the predictions of traffic density (top row) and traffic velocity (bottom row). Fig.~\ref{fig:ngsim_result_prediction}(a) and Fig.~\ref{fig:ngsim_result_prediction}(d) present the heatmaps of traffic density and traffic velocity in spatio-temporal space. Those two predictions are close to the ground truth shown in Fig.~\ref{fig:ngsim_data}. The other 4 subfigures show the snapshots of the prediction intervals of the traffic density and velocity.

\begin{figure}[h!]
  \centering
  \includegraphics[width=\linewidth]{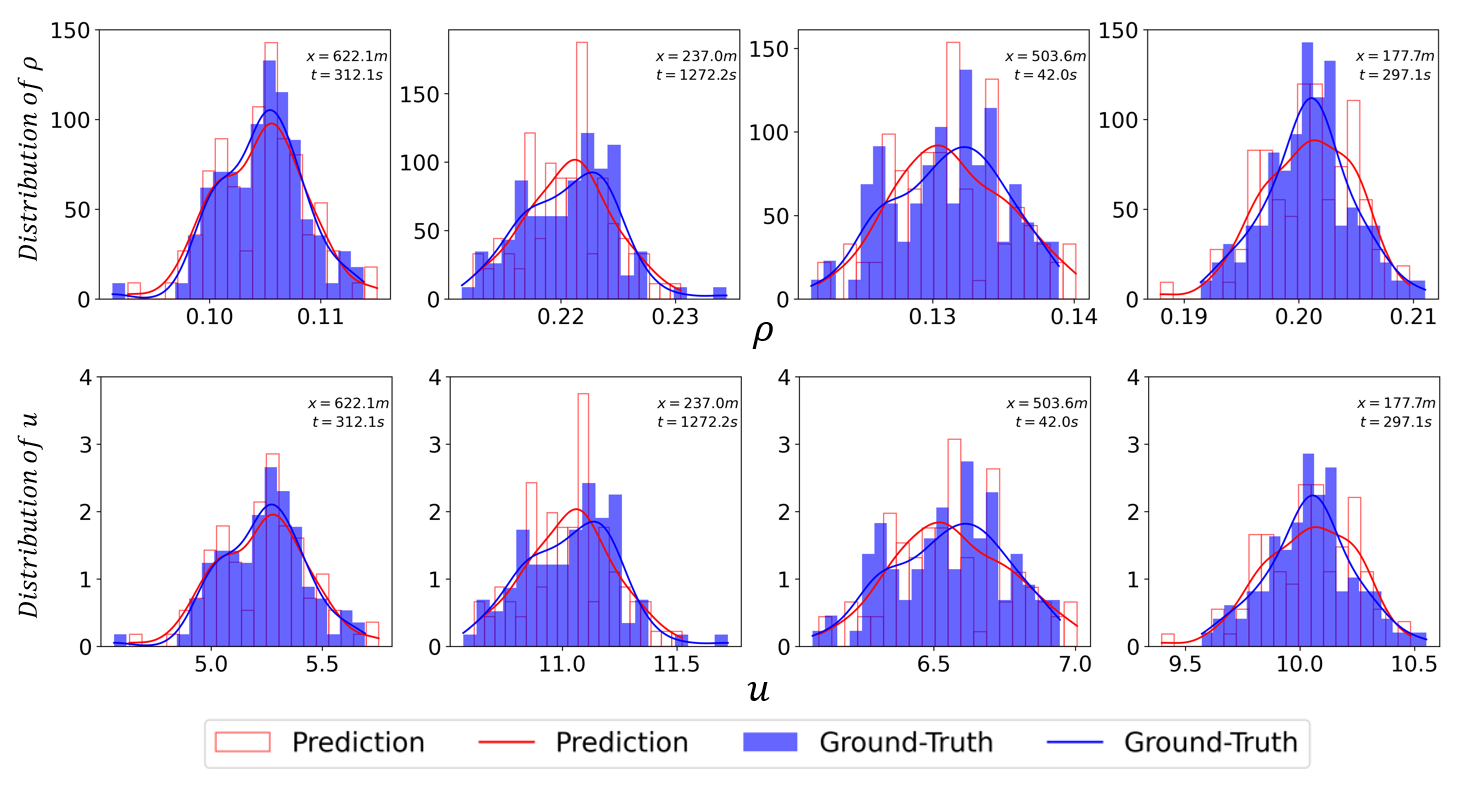}
  \caption{Prediction distributions of TrafficFlowGAN for the NGSIM dataset.}
  \label{fig:ngsim_result_viz}
\end{figure}

Fig.~\ref{fig:ngsim_result_viz} presents the comparison between the ground-truth traffic density distribution and that predicted by the TrafficFlowGAN model, each subfigure for a randomly sampled spatio-temporal coordinates. Most parts of the predicted and ground-truth distributions overlap with each other, which demonstrates that our proposed model can estimate the real-world traffic states uncertainties well.

\begin{figure}[h!]%
    \centering
    \subfloat[\centering ]{{\includegraphics[height=0.23\columnwidth]{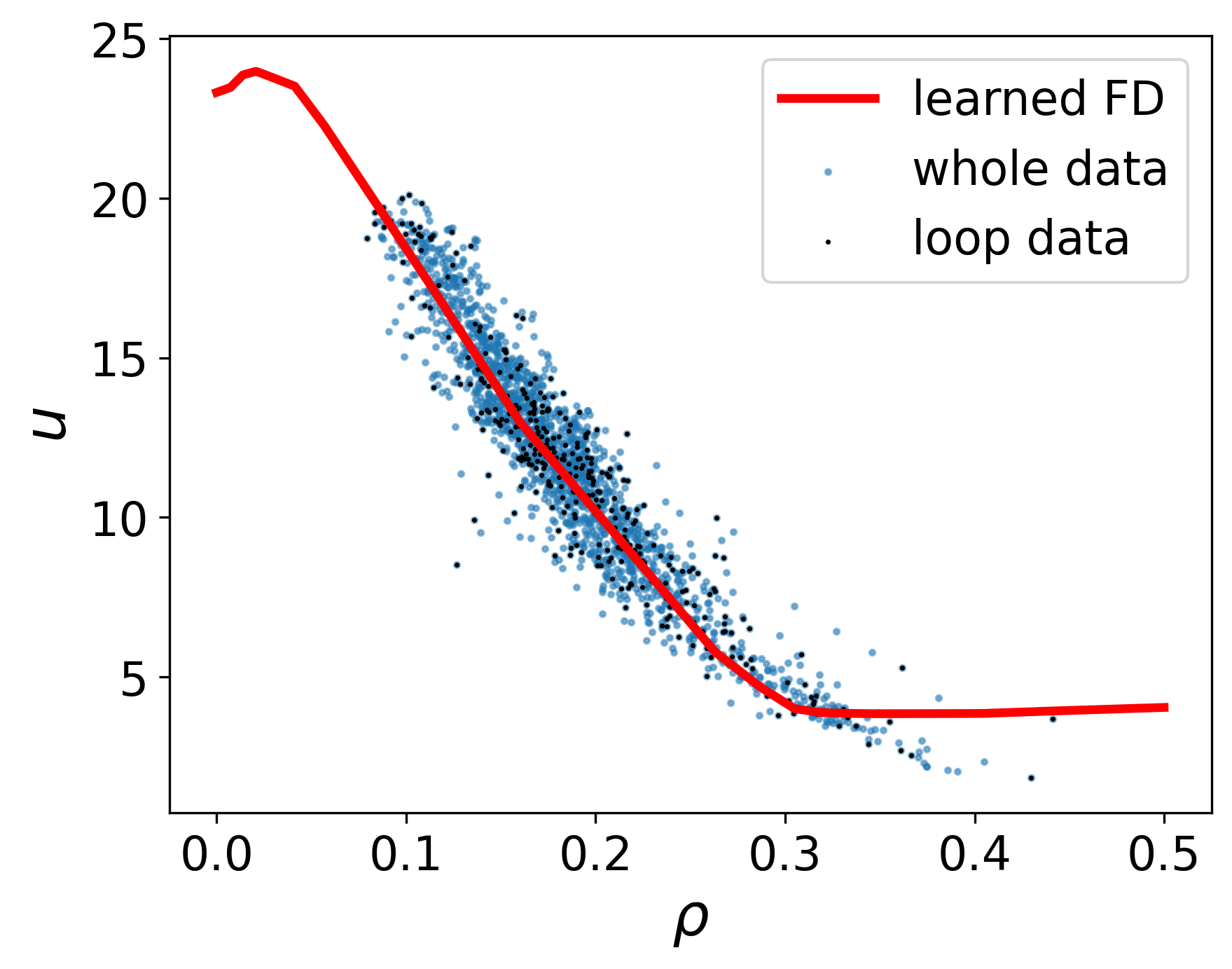} }}\hspace{0mm}
    \subfloat[\centering ]{{\includegraphics[height=0.23\columnwidth]{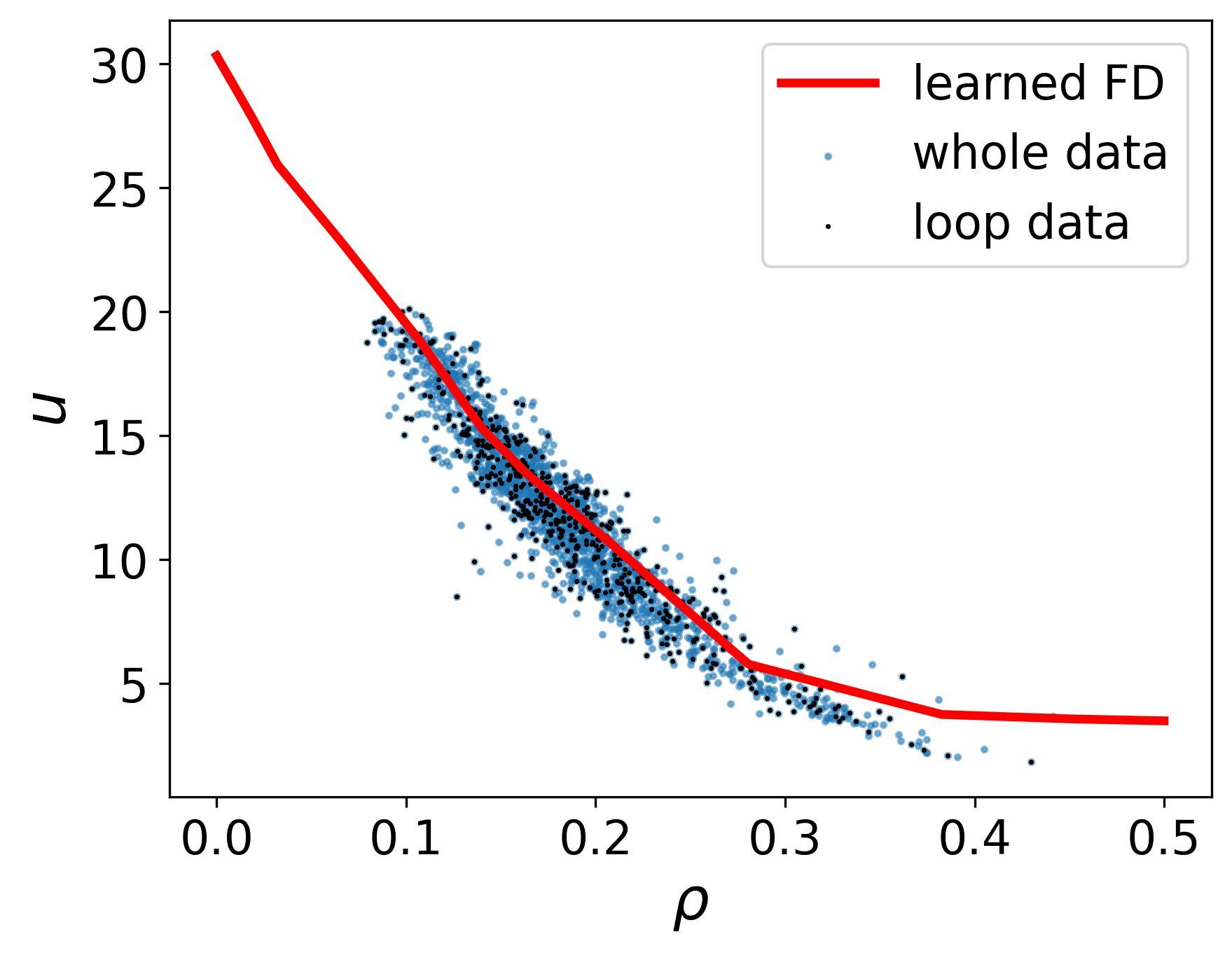} }}\hspace{0mm}
     \subfloat[\centering ]{{\includegraphics[height=0.23\columnwidth]{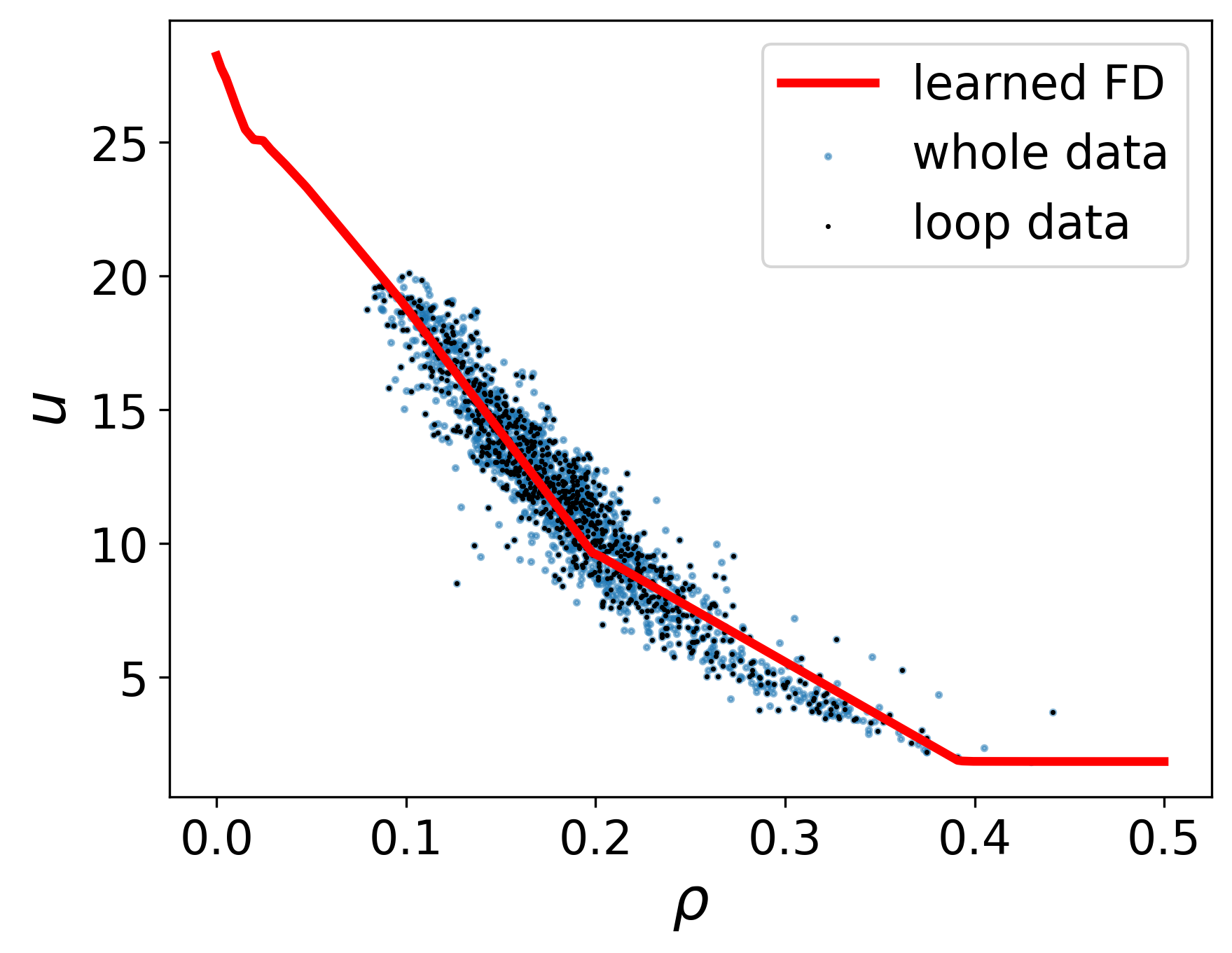} }}\hspace{0mm}
    \caption{ Traffic density-velocity relations learned by the surrogate model (s-net) for different loop detector numbers. (a) loop detector number is 4. (b) loop detector number is 6. (c) loop detector number is 10.}%
    \label{fig:ngsim_result_fd}%
\end{figure}

Fig.~\ref{fig:ngsim_result_fd} shows the learned traffic density-velocity relation by TrafficFlowGAN. When the number of loop detectors is 4, TrafficFlowGAN can already capture this relation well. Increasing the number of loop detectors helps TrafficFlowGAN learn a subtler pattern. 

\vspace{-2mm}
\section{Conclusions}\label{sec:conclusion}
\vspace{-2mm}
This paper proposes TrafficFlowGAN to quantify the uncertainty in dynamical systems. TrafficFlowGAN leverages MLE, adversarial training, and PIDL to generate high-quality samples with exact data likelihood and efficient data usage. To verify that TrafficFlowGAN can learn the solutions of a second-order PDE, we conduct a numerical experiment where data is generated from a known ARZ equation. Numerical results show that TrafficFlowGAN manages to reconstruct the PDE solutions and recover the underlying relation between state variables. We further apply TrafficFlowGAN to a TSE problem using real-world data to demonstrate its performance. Results show that TrafficFlowGAN can better capture the real-world uncertainty than baselines, including the pure flow, the physics-informed flow, and the flow based GAN. We also show that TrafficFlowGAN can learn the real-world traffic density-velocity relation simultaneously.

This work can be further improved in two directions. First, apart from the weighted sum, other approaches to integrating the likelihood loss, adversarial loss, and physics loss can be proposed. Second, TrafficFlowGAN needs to be re-trained if applied to
other roads or to the same road but within a new time
slot. We will work on the generalizability of TrafficFlowGAN in the future.



%
%

\bibliographystyle{splncs04}

\bibliography{acmart.bib}
\end{document}


\title{Supplementary}

\author{}
\institute{}
\maketitle

\section{Discriminator}

For the discriminator, we use different configurations for different numbers of loop detectors. Table.~\ref{tab:cnn} shows the configurations of the discriminator under different number of loop detectors. Conv2D, Flatten and FC denote 2-D convolution layer, flatten layer and fully connected layers, respectively. We use batch normalization after each convolution layer. The activation function after each layer is the hyperbolic tangent activation function (Tanh)\footnote{code is available at: \href{https://anonymous.4open.science/r/TrafficFlowGAN}{\tcb{https://anonymous.4open.science/r/TrafficFlowGAN}}}.

\begin{table*}
\centering
\renewcommand{\arraystretch}{1.2}
\begin{tabular}{ |p{2cm}||p{2.5cm}|p{2.5cm}|p{2.5cm}|  }
 \hline
 \multicolumn{4}{|c|}{Discriminator network structure} \\
 \hline
 Layers& $\#$loop=3 &$\#$loop=4&$\#$loop=6\\
 \hline
 Conv2D   &   $3\times1\times4$    &$4\times2\times4$   &  $3\times2\times4$\\
 Conv2D   &   $3\times1\times8$    &$4\times2\times8$   &  $3\times2\times8$\\
 Conv2D   &   $3\times1\times16$   &$4\times2\times16$  &  $3\times2\times16$\\
 \hline
 Flatten          &  &  &\\
 \hline
 FC       &   144                  & 144                &  240\\
 FC       &   64                   & 64                 &  64\\
 
 \hline
 \hline
 Layers& $\#$loop=10 &$\#$loop=14&$\#$loop=18\\
  \hline
 Conv2D   &   $3\times2\times4$    &$3\times2\times4$   &  $3\times2\times4$\\
 Conv2D   &   $3\times2\times8$    &$3\times2\times8$   &  $3\times2\times8$\\
 Conv2D   &   $3\times2\times12$   &$3\times2\times12$  &  $3\times2\times12$\\
 \hline
 Flatten          &  &  &\\
 \hline
 FC       &   96                  & 96                &  144\\
 FC       &   64                   & 64                 &  64\\
 
 \hline
\end{tabular}
\caption{Network structures for the discriminator under different loop detectors. For Conv2D layers, the first two numbers indicate the kernel sizes, and the last number indicates the number of output channels. For the FC layer, the number indicates the number of hidden neurons.}
\label{tab:cnn}
\end{table*}

\section{Generator}

For our conditional flow model, the prior network(p-net) consists of networks for the prior mean $\pmb{\mu}$ and networks for the prior standard deviation $\pmb{\sigma}$. Each of these two prior networks has 6 affine coupling layers, and each of the layers is a fully connected layer with 256 neurons. We use Leaky Rectified Linear Unit(Leaky ReLU) as the activation function.

Each affine coupling layer in the generator consists of a scale function (k-net) and a translation function (b-net). In the experiments, we use 8 affine coupling layers for k-net and b-net, respectively. Every single layer is a fully connected layer with 256 neurons. The activation function after each layer is Rectified Linear Unit (ReLU). We use batch normalization in our experiments.